\documentclass[runningheads]{llncs}

\usepackage{eccv}

\usepackage{eccvabbrv}

\usepackage{graphicx}
\usepackage{booktabs}
\usepackage{multirow}

\usepackage[accsupp]{axessibility}  

\usepackage{hyperref}

\usepackage{orcidlink}

\usepackage{longtable}
\usepackage{tabularx}
\usepackage{array}
\usepackage{tcolorbox}
\usepackage{listings}

\begin{document}

\title{Anatomy Contextualized Adaptation of CT Foundation Models} 

\titlerunning{Anatomy Contextualized Adaptation}

\author{Roshan Kenia\inst{1, 2},
Stephanie L McNamara\inst{3}, \and
William Lotter\inst{2, 4, 5}}

\authorrunning{R.~Kenia et al.}

\institute{Department of Biomedical Informatics, Harvard Medical School, Boston, MA \and
Department of Data Science, Dana-Farber Cancer Institute, Boston, MA \and 
Department of Radiology, Massachusetts General Hospital, Boston, MA \and
Department of Pathology, Brigham and Women’s Hospital, Boston, MA \and
Department of Pathology, Harvard Medical School, Boston, MA}

\maketitle

\begin{abstract}
CT vision-language foundation models have demonstrated promising performance across downstream tasks, but are typically trained with whole-volume representations that dilute fine-grained anatomical signals. Fine-grained vision-language pre-training addresses this by aligning anatomy-level visual features with anatomy-specific text, but in doing so discards the global context that whole-volume models provide. Furthermore, existing fine-grained approaches train from scratch, making them computationally expensive. We introduce Anatomy Contextualized Adaptation (ACA), a lightweight framework that adapts frozen CT foundation model representations for anatomy-level vision-language alignment while enhancing global contextualization. ACA uses TotalSegmentator to decompose CT volumes into anatomy-level embeddings, which are refined via a transformer that captures cross-anatomy relationships, and aligned to both per-anatomy and scan-level text extracted from radiology reports. Evaluated on Merlin and CT-RATE, ACA consistently outperforms both the frozen foundation model baselines and existing fine-grained methods in zero-shot finding classification, while requiring less than one hour of training once embeddings are cached. The attention weights learned by ACA's inter-anatomy transformer additionally indicate plausible cross-anatomy context routing. Altogether, these results support ACA as a lightweight approach for adapting CT foundation models to anatomically grounded vision-language alignment while preserving and enhancing global anatomical context\footnote{Code is available at \href{https://github.com/lotterlab/ACA}{https://github.com/lotterlab/ACA}}.
\end{abstract}

\section{Introduction}
\label{sec:intro}

CT foundation models \cite{zhou2021models, xie2022unimiss, wu2025towards, wu2024voco} trained on large-scale data have demonstrated promising performance across a wide range of downstream tasks, including zero-shot disease classification via vision-language contrastive learning. However, these models are typically trained using whole-volume CT representations, condensing a 3D volume into a single embedding for pre-text training \cite{merlin, ct_clip, percival}. Conversely, fine-grained vision-language pre-training (FVLP) has emerged as a strategy to align anatomy-specific visual embeddings with corresponding textual descriptions \cite{fvlm, visd_boost, ct_glip, zhang2026gcl}. These methods can recover fine-grained signal, but in doing so discard the global context that whole-volume models provide. Furthermore, existing FVLP approaches have relied on training the entire model from scratch, which is computationally expensive and can be difficult to scale.

We introduce \textit{Anatomy Contextualized Adaptation} (ACA), a framework for adapting pretrained CT foundation models to anatomy-level vision-language alignment while enhancing global context. ACA combines the strengths of both whole-volume foundation models and FVLP: the broad, frozen representations of pretrained CT foundation models are adapted using lightweight trainable modules to produce anatomy-level embeddings, which are then contextualized across anatomies using transformer-based attention. The training loss includes both anatomy-level and scan-level vision-language alignment. Across two datasets and two base foundation models, ACA boosts zero-shot diagnostic performance and its learned attention indicates plausible cross-anatomy context routing.

\section{Related Work}

\textbf{CT Foundation Models.} Recent work has explored large-scale pre-training of CT models that can be adapted to a variety of clinical tasks. Vision-language pre-training approaches, building on image-text contrastive learning \cite{radford2021learning, wang2022medclip, wu2023medklip}, align CT volumes with paired radiology reports \cite{percival, cao2024bootstrapping}. CT-CLIP \cite{ct_clip} and Merlin \cite{merlin} represent the dominant paradigm, training contrastively on chest/abdominal CT volumes paired with free-text reports, enabling zero-shot abnormality detection at supervised-level performance. Vision-only approaches such as CT-FM \cite{ct_fm} and 3DINO \cite{3dino} leverage self-supervised contrastive \cite{chen2020simple, he2020momentum, caron2021emerging} and masked image modeling \cite{bao2021beit, he2022masked, xie2022simmim} objectives on unlabeled CT volumes, demonstrating strong transfer to segmentation and classification tasks. Segmentation-oriented foundation models such as VISTA3D \cite{vista3d} and SAM-Med3D \cite{wang2025sam}, together with large-scale anatomical segmentation frameworks such as TotalSegmentator \cite{wasserthal2023totalsegmentator}, enable generalized anatomical understanding across dozens to hundreds of anatomical structures in 3D CT imaging. Task-specific multimodal models including M3FM \cite{m3fm} and LCTfound \cite{lct_found} incorporate structured clinical data and imaging to address lung cancer screening and other downstream workflows. Given the emergent zero-shot capabilities of vision-language approaches, we focus on adapting Merlin and CT-CLIP while leveraging TotalSegmentator to facilitate fine-grained modeling.

\begin{figure}[t]
    \centering
    \includegraphics[width=\linewidth]{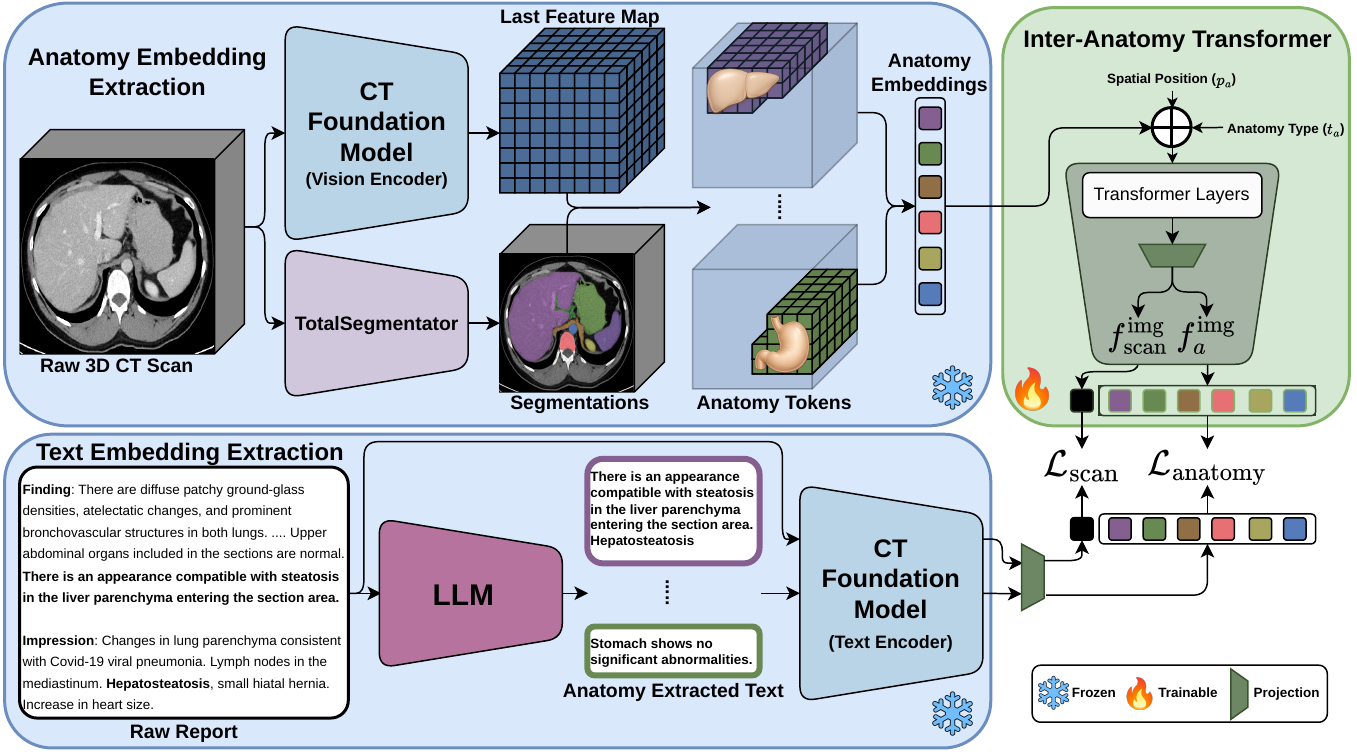}
    \caption{
    Overview of ACA. Anatomy-level visual embeddings are constructed from a frozen CT foundation model using TotalSegmentator segmentations, while anatomy-level text embeddings are extracted from radiology reports using an LLM and the corresponding frozen text encoder. An inter-anatomy transformer, augmented with spatial position and anatomy type embeddings, contextualizes the anatomy embeddings across the full set of structures. The resulting embeddings are aligned to per-anatomy text via $\mathcal{L}_{\text{anatomy}}$, while their mean-pooled scan-level embedding is aligned to the full report via $\mathcal{L}_{\text{scan}}$.
    }
    \label{fig:methodology}
\end{figure}

\textbf{Fine-grained Vision-language Pre-training.} Rather than aligning entire CT volumes with full radiology reports, fine-grained vision-language pretraining (FVLP) aligns individual anatomical regions with their corresponding report descriptions \cite{huang2021gloria, fvlm, muller2022joint, wang2022multi, zhang2026gcl}. fVLM \cite{fvlm} explicitly decomposes CT volumes using TotalSegmentator \cite{wasserthal2023totalsegmentator} and matches anatomy-level visual tokens to anatomy-specific text via cross-attention. ViSD-Boost \cite{visd_boost} identifies a semantic density gap between low signal-to-noise visual representations and information-dense diagnostic reports, and addresses it through disease-level visual contrastive learning and anatomical normality modeling.

Despite these advances, existing FVLP methods treat each anatomy independently during both training and inference, ignoring the inter-anatomy context that is often essential for accurate diagnosis \cite{wen2024biological}. For example, organomegaly (abnormal enlargement of an organ; e.g., hepatomegaly, splenomegaly) is best assessed relative to surrounding structures and body habitus. Our work addresses this gap by introducing an inter-anatomy transformer that contextualizes anatomy embeddings across all present anatomies before alignment, enabling richer and more clinically grounded representations.

\textbf{CT Foundation Model Datasets.} The datasets used to train the Merlin and CT-CLIP foundation models have been publicly released. The Merlin \cite{merlin} dataset is a large-scale abdominal CT cohort comprising 25,494 CT scans paired with radiology reports from 18,317 unique patients, collected at Stanford University Medical Center, with annotations spanning 30 abdominal findings. CT-CLIP was developed using CT-RATE \cite{ct_clip}, a chest CT dataset comprising 50,188 reconstructed 3D volumes from 25,692 scans of 21,304 unique patients, each paired with a radiology report and 18 annotated abnormality labels extracted via an automated text classifier.

\section{Methodology: Anatomy Contextualized Adaptation}

ACA is illustrated in Figure~\ref{fig:methodology} and consists of three core components. First, anatomy-level visual embeddings are extracted from a frozen CT foundation model by pooling its feature maps based on TotalSegmentator segmentations, while anatomy-level text embeddings are extracted from the corresponding radiology report using an LLM. Second, an inter-anatomy transformer, augmented with learned spatial position and anatomy type embeddings, contextualizes each anatomy's embedding using the full set of structures present in the scan. Finally, the contextualized anatomy embeddings are projected into a shared contrastive space and aligned to their corresponding per-anatomy text, while a mean-pooled scan-level embedding is aligned to the full report embedding to provide a complementary global supervision signal. We describe each component in detail below.

\subsection{Embedding Construction}
\label{sec:extraction}

\textbf {Anatomy Embedding Extraction.} To construct anatomy-specific representations, we first apply TotalSegmentator \cite{wasserthal2023totalsegmentator} to segment each CT volume into 44 anatomical structures. The 44 structures are based on condensing TotalSegmentator's original 117 classes into a set of related anatomical groups (e.g., left kidney and right kidney $\rightarrow$ kidney), as summarized in Appendix Table~\ref{anatomy_mapping}.

For each CT volume, we pass the scan through a frozen foundation model backbone to obtain intermediate spatial feature representations. We experiment with two foundation models: Merlin, which uses an I3D ResNet152 \cite{carreira2017quo} encoder, and CT-CLIP, which uses a CT-ViT \cite{ct_vit} encoder. Merlin uses a $224 \times 224 \times 160$ voxel input, and generates a final feature map of size $2048 \times 10 \times 7 \times 7$. CT-CLIP uses a $480 \times 480 \times 240$ voxel input, and produces a feature map of size $512 \times 24 \times 24 \times 24$. Both models subsequently pool these features to generate a scan-level representation for downstream tasks, whereas we construct anatomy-level features from them.

To obtain anatomy-level representations, we project the TotalSegmentator segmentation masks into the spatial resolution of the extracted feature maps. Specifically, we apply a non-overlapping max-pool over each binary organ mask using a kernel matched to the backbone's effective patch size $(32 \times 32 \times 16)$ voxels for Merlin and $(40 \times 40 \times 20)$ voxels for CT-CLIP), yielding a discrete patch-presence grid at the feature map resolution. For each of the 44 anatomical structures, we extract all features in which the structure occupies a patch across the last feature map of the backbone. We subsequently mean-pool these features into a single structure embedding. Structures for which TotalSegmentator produces no non-empty segmentation mask in a given scan are considered absent and excluded from that scan's input sequence. 
For each anatomy $a$ with $N_a$ final layer extracted features $\{v_1, \ldots, v_{N_a}\} \in \mathbb{R}^d$, we denote the anatomy embedding as $
\bar{v}_a = \frac{1}{N_a} \sum_{i=1}^{N_a} v_i
$, where $d = 2048$ for the Merlin backbone and $d = 512$ for CT-CLIP. 

Each embedding is subsequently $\ell_2$-normalized. In addition, we compute inter-anatomy spatial position encodings for each anatomical structure as a 44-dimensional vector, where the $g$-th entry is the $\ell_2$ distance between the structure's voxel-space centroid and the centroid of structure $g$, normalized by the volume diagonal length $\sqrt{D^2 + H^2 + W^2}$, where $D$, $H$, $W$ correspond to the number of voxels along the depth, height, and width, respectively. These encodings capture the geometric relationships between anatomical structures within a volume and are used as positional signals in downstream models.

\textbf{Text Embedding Extraction.} For the text modality, anatomy-specific findings are extracted from radiology reports using Qwen3-4B-Instruct~\cite{yang2025qwen3} (LLM in Figure \ref{fig:methodology}), following a similar strategy to \cite{fvlm}. These findings are subsequently encoded using the corresponding frozen Merlin or CT-CLIP text encoder to produce anatomy-level text embeddings. Both the prompts used for anatomy-specific finding extraction and an example of the resulting anatomy-specific findings are shown in Appendix Figures ~\ref{qwen_prompt} and ~\ref{fig:anatomy_text_examples}, respectively. A report-level text embedding is also generated by passing the full report through the text encoder.

\subsection{Inter-Anatomy Transformer}
\label{sec:ACA}

To enable contextual reasoning across anatomical structures within a scan, we pass all present anatomy embeddings jointly through a transformer encoder. Prior to transformer input, each anatomy embedding is projected and combined with two learned tokens. The first consists of an MLP applied to a positional encoding vector $p_a \in \mathbb{R}^{44}$, which encodes the normalized distances from anatomy $a$ to all other structures in the taxonomy as described in Section \ref{sec:extraction}. The second token is a learnable anatomy type embedding $t_a$ specific to each of the 44 anatomy groups. The final anatomy embedding for transformer input, $h_a$, is computed as follows:

\begin{equation}
h_a = \bar{v}_a W_{\text{vis}} + \text{MLP}(p_a) + t_a
\end{equation}

The sequence $\{h_a\}_{a \in \mathcal{P}}$, where $\mathcal{P}$ denotes the set of anatomical structures present in the scan, is passed through an $L$-layer transformer encoder with pre-layer normalization, producing a contextualized embedding for each anatomy.

\subsection{Combined Anatomy-Level and Global Report Loss}
\label{sec:loss}

We supervise the inter-anatomy transformer with two complementary loss terms operating at different granularities: an anatomy-level contrastive loss, $\mathcal{L}_{\text{anatomy}}$, that aligns each anatomy's contextualized embedding to its corresponding text description, and a scan-level report loss, $\mathcal{L}_{\text{scan}}$, that grounds the full anatomy sequence to the global radiology report.

To enable the contrastive losses, the anatomy-level image and text embeddings are first projected to a shared $d'$-dimensional contrastive space. For both embedding types, a two-layer projection head is used, consisting of a linear layer, GELU activation, layer normalization, a second linear layer, and $\ell_2$ normalization in sequence. We denote the final anatomy-level image and text embeddings as $f_a^{\text{img}}$ and $f_a^{\text{txt}}$ respectively. To enable the scan-level loss, a scan-level image embedding is computed as the $\ell_2$-normalized mean over anatomy embeddings: $f_{\text{scan}}^{\text{img}} = \ell_2\!\left(\frac{1}{|\mathcal{P}|}\sum_{a \in \mathcal{P}} f_a^{\text{img}}\right)$. The scan-level text embedding $f_{\text{scan}}^{\text{txt}}$ is obtained by passing the full radiology report through the frozen text encoder and then projecting to the shared contrastive space using the same text projection head as the per-anatomy text embeddings.

\textbf {Anatomy-Level Loss.}
For each anatomical structure $s$, we collect the $N_s$ instances present across the batch and compute a symmetric image-text contrastive loss over them. The standard contrastive loss uses a hard diagonal objective wherein matching image-text pairs are treated as positives and all other combinations are treated as negatives. To handle the prevalence of false negatives in this formulation, where anatomical structures from different patients may be semantically identical yet penalized as different, we follow \cite{fvlm} and construct a soft target matrix $T \in [0,1]^{N_s \times N_s}$, where $i$ and $j$ index instances of structure $s$:

\begin{equation}
T_{ij} = \mathbf{1}[i = j]
       + \mathbf{1}[\text{normal}(i) \wedge \text{normal}(j)]
       + \mathbf{1}[\text{patient}(i) = \text{patient}(j),\; i \neq j]
\end{equation}

where $\text{normal}(\cdot)$ are binary normality labels extracted from radiology reports using Qwen3-4B-Instruct (see Figure~\ref{fig:prompt_templates}), and each row is normalized to sum to one. Anatomies where all instances are normal are skipped entirely, focusing capacity on pathological variation. The per-anatomy loss is then:

\begin{equation}
\mathcal{L}_{s} = -\frac{1}{2N_s} \sum_{i=1}^{N_s} \sum_{j=1}^{N_s} T_{ij} \left[ \log \frac{e^{f_i^{\mathrm{img}} \cdot f_j^{\mathrm{txt}} / \tau}}{\sum_k e^{f_i^{\mathrm{img}} \cdot f_k^{\mathrm{txt}} / \tau}} + \log \frac{e^{f_i^{\mathrm{txt}} \cdot f_j^{\mathrm{img}} / \tau}}{\sum_k e^{f_i^{\mathrm{txt}} \cdot f_k^{\mathrm{img}} / \tau}} \right]
\end{equation}

where $f_i^{\mathrm{img}}$ and $f_i^{\mathrm{txt}}$ are the projected and $\ell_2$-normalized visual and text embeddings for instance $i$, and $\tau$ is a learnable temperature initialized at $0.07$ and clamped to $[0.001, 0.5]$. As seen in Equation \ref{l_anatomy}, the total anatomy-level loss, $\mathcal{L}_{\text{anatomy}}$, sums over all active structures $\mathcal{S}$, defined as those that appear in at least one scan in the batch and have at least one abnormal instance. Scans with no present anatomies are excluded from all loss terms.

\begin{equation}
\label{l_anatomy}
\mathcal{L}_{\text{anatomy}} = \sum_{s \in \mathcal{S}} \mathcal{L}_s
\end{equation}

\textbf {Scan-Level Report Loss.}
While $\mathcal{L}_{\text{anatomy}}$ aligns each anatomy independently to short descriptive text, it provides no signal connecting the anatomy sequence as a whole to the broader clinical context of the scan. To bridge this gap, we include a scan-level contrastive loss, $\mathcal{L}_{\text{scan}}$, that aligns the scan-level visual embedding, $f_{\text{scan}}^{\text{img}}$, to the text embedding of the full radiology report, $f_{\text{scan}}^{\text{text}}$. Unlike the anatomy-level loss, each scan is matched only to its own report, so a hard diagonal objective is used:

\begin{equation}
\mathcal{L}_{\text{scan}}
= -\frac{1}{2B}
\sum_{i=1}^{B}
\left[
\log \frac{e^{f_{\text{scan},i}^{\text{img}} \cdot f_{\text{scan},i}^{\text{txt}} / \tau_{\text{scan}}}}
         {\sum_{k=1}^{B} e^{f_{\text{scan},i}^{\text{img}} \cdot f_{\text{scan},k}^{\text{txt}} / \tau_{\text{scan}}}}
+
\log \frac{e^{f_{\text{scan},i}^{\text{txt}} \cdot f_{\text{scan},i}^{\text{img}} / \tau_{\text{scan}}}}
         {\sum_{k=1}^{B} e^{f_{\text{scan},i}^{\text{txt}} \cdot f_{\text{scan},k}^{\text{img}} / \tau_{\text{scan}}}}
\right]
\end{equation}

where $B$ is the number of scans in the batch with at least one present anatomy, and $\tau_{\text{scan}}$ is a separate learnable temperature initialized at $0.07$ and clamped to $[0.001, 0.5]$. Because $f_{\text{scan}}^{\text{img}}$ is a mean over the same $f_a^{\text{img}}$ embeddings optimized by $\mathcal{L}_{\text{anatomy}}$, the scan-level signal propagates directly through the visual projection head, jointly shaping a space where individual embeddings are discriminative at the anatomy level and coherent at the scan level.

\textbf {Total Loss.}
The combined objective is:

\begin{equation}
\mathcal{L}_{\text{total}} = \mathcal{L}_{\text{anatomy}} + \lambda\mathcal{L}_{\text{scan}}
\end{equation}

We refer to this full model as \textbf{ACA}. We use $\lambda=1$ in our experiments to provide a simple balance of local and global signals, while also performing two ablations: \textbf{ACA w/o $\mathcal{L}_{\text{scan}}$}, which uses only $\mathcal{L}_{\text{anatomy}}$, and \textbf{ACA w/o $\mathcal{L}_{\text{anatomy}}$}, which uses only $\mathcal{L}_{\text{scan}}$.

\begin{figure}[t]
    \centering
    \includegraphics[width=\linewidth]{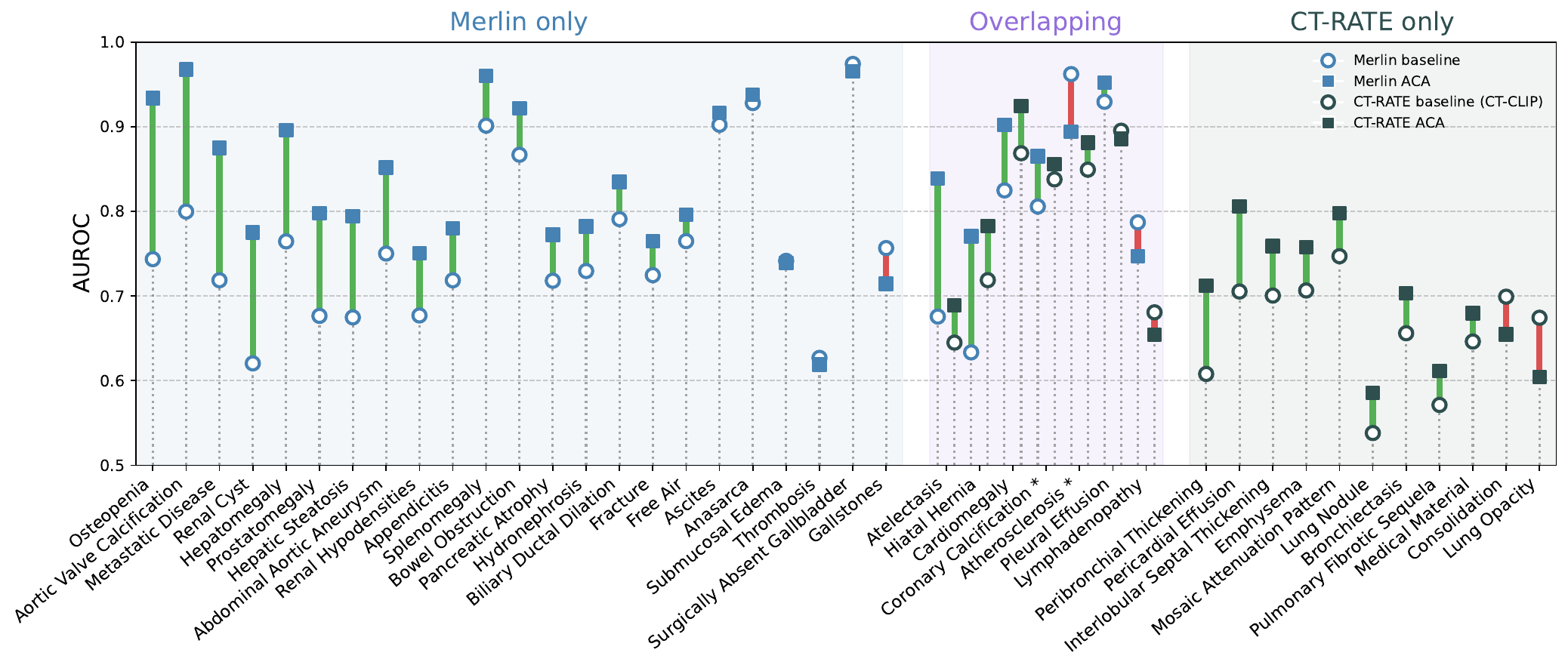}
    \caption{
    Per-finding AUROC comparison between the original models and the ACA anatomy-guided model for In-Distribution zero-shot evaluation. \textcolor{Green}{Green} lines represent improvement over the baseline, \textcolor{Red}{Red} lines represent regression. * These findings' names from CT-RATE are simplified to match with Merlin: Coronary Artery Wall Calcification = Coronary Calcification, Arterial Wall Calcification = Atherosclerosis.
    }
    \label{fig:per_finding_auroc}
\end{figure}

\section{Experiments}
\subsection{Datasets and Training Setup}

We train on two CT datasets using their respective frozen foundation model backbones. For Merlin \cite{merlin}, we use the creator-defined splits of 15,314 training, 5,055 validation, and 5,125 test scans. For CT-RATE \cite{ct_clip}, we partition the training data into 37,545 training and 9,598 validation scans, and evaluate on the 3,039-scan test set defined by the dataset creators. For each dataset, we train ACA and each baseline independently using AdamW with learning rate $1 \times 10^{-4}$, batch size 64, and 50 epochs, selecting the best model for evaluation based on validation loss. Full hyperparameters are reported in Appendix Tables~\ref{tab:shared_hparams} and~\ref{tab:arch_hparams}.

\subsection{Evaluation}
\label{sec:eval}
We evaluate each model on the held-out test split defined by the organizers of each dataset. We focus on zero-shot finding classification, where the findings present in each dataset are detailed in Appendix Tables \ref{tab:merlin-findings} and \ref{tab:ctrate-findings}. Each finding is treated as binary classification (positive or negative) at the scan-level. Unlike Merlin \cite{merlin}, we do not subsample negative samples to match the number of positives and evaluate on all samples for a finding.

To evaluate similarity with the model's scan-level visual embedding, we associate each finding with a set of positive prompts describing the pathology and negative prompts describing normal appearance (see Appendix Figure~\ref{fig:finding_prompts}). These are encoded by the frozen foundation model text encoder used by the model, averaged within each class (positive or negative), and projected through the model's trained text projection head to obtain $f^{+}$ and $f^{-} \in \mathbb{R}^{d'}$. The model's scan-level score for the finding (defined in Appendix \ref{supp:anatomy_guided}) is computed as the difference in cosine similarity between the model's scan-level image representation ($f_{\text{scan}}^{\text{img}}$) and each class embedding ($f^{+}$ and $f^{-}$), with all embeddings $\ell_2$ normalized. In the case where no structures are detected, the scan embedding defaults to a zero vector, which scores 0 against all finding prompts and predicts negative. Further details are provided in Appendix~\ref{supp:anatomy_guided}.

We evaluate each model under in-distribution and out-of-distribution settings. In the in-distribution setting, models are trained and tested on the same dataset using the corresponding foundation model (e.g., Merlin train $\to$ Merlin test). Out-of-distribution tests generalization across datasets and base foundation model (e.g., Merlin train $\to$ CT-RATE test). As the Merlin and CT-RATE datasets cover different anatomies and finding sets, out-of-distribution evaluation is restricted to the 7 findings shared between the datasets (see Figure \ref{fig:per_finding_auroc}).

\subsection{Baselines}
\label{sec:baselines}

We compare to several baselines to highlight the effects of each modeling aspect of ACA. Each baseline uses the same prompting strategy to generate positive and negative text embeddings for each finding, using the corresponding text encoder associated with the model. 

\textbf {Global VLMs.} We assess the baseline performance of Merlin and CT-CLIP, using the original, fixed scan-level vision embeddings for zero-shot similarity.

\textbf {Fine-grained adaptation.} To isolate the effect of fine-grained adaptation on top of frozen foundation model embeddings, we evaluate an \textbf{MLP} baseline that does not include the inter-anatomy transformer in ACA. In this baseline, each anatomy's pre-computed mean embedding is passed directly through a two-layer projection head consisting of a linear layer, GELU activation, layer normalization, and a second linear layer, with $\ell_2$-normalization applied to the output. No inter-anatomy context is used. Each anatomy is projected independently and the model is trained with the $\mathcal{L}_{\text{anatomy}}$ loss alone.

We additionally compare to two existing fine-grained architectures, using these approaches to generate anatomy-level embeddings on top of the foundation model features. This adaptation enables a compute-matched comparison, but means these baselines do not necessarily reflect the performance of the originally published methods which rely on end-to-end training (see Limitations). The first is the anatomy query pooling mechanism from \textbf{fVLM} \cite{fvlm}. For the Merlin backbone, features from all four I3D ResNet152 layers are concatenated to form $3840$-dimensional multi-scale representations per patch. For the CT-CLIP CT-ViT backbone, features from the last layer of the spatial transformer and the last layer of the causal transformer are concatenated to form $1024$-dimensional representations. A learned anatomy-specific query vector then attends over these features via cross-attention to produce a single anatomy embedding, which is projected and aligned to the corresponding per-anatomy text using $\mathcal{L}_{\text{anatomy}}$.

 The second architecture is an adaptation of \textbf{ViSD-Boost} \cite{visd_boost}, which amplifies disease signals by modeling the normal distribution of each anatomy in latent space via a VQ-VAE. Because our framework operates on frozen embeddings, we adapt the method to three training stages for fair comparison. First, we perform anatomy-level contrastive alignment using the fVLM query pooling mechanism, with normal--normal anatomy pairs upweighted in the contrastive targets, rather than a vision-only pre-training stage. Second, we train the Transformer-based VQ-VAE exclusively on normal anatomy instances to learn a shared anatomy-conditioned codebook of healthy representations. Finally, with the VQ-VAE frozen, the original multi-scale anatomy tokens are fused with their VQ-VAE reconstruction through a residual projection, with the hybrid embedding aligned to per-anatomy text as other fine-grained baselines via $\mathcal{L}_{\text{anatomy}}$ rather than binary positive/negative prompts. The disease-level contrastive pre-training stage and momentum encoder from the original method are omitted as they require end-to-end vision encoder training.

\textbf {Global adaptation.} To isolate the effect of anatomy-level, fine-grained training in ACA, we define a \textbf{Spatial Transformer} baseline that bypasses anatomy segmentation entirely, but includes a module analogous to the inter-anatomy transformer in ACA. In this baseline, the final feature map of the frozen backbone is depth mean-pooled and reshaped into a sequence of spatial patch features, which are processed by a transformer encoder. The mean of all spatial patch outputs is then projected through a two-layer projection head and used as the scan-level image representation, followed by radiology report alignment using $\mathcal{L}_{\text{scan}}$.

\begin{table}[t]
\centering
\caption{AUROC comparison across MERLIN and CT-RATE datasets. $^\dagger$ These methods were adapted to operate on frozen embeddings. * Denotes out-of-distribution evaluation restricted to the 7 shared findings, where models trained on CT-RATE are evaluated on Merlin* and models trained on Merlin are evaluated on CT-RATE*. Per finding results can be found in Appendix Tables \ref{tab:merlin_id}, \ref{tab:ctrate_id}, \ref{tab:merlin_ood}, and \ref{tab:ctrate_ood}.}
\label{tab:auroc_comparison}
\resizebox{\textwidth}{!}{%
\begin{tabular}{lcc|cc|c}
\hline
& \multicolumn{2}{c|}{In-distribution} &
\multicolumn{2}{c}{Out-of-distribution} \\
Model & Merlin & CT-RATE & Merlin* & CT-RATE* & Average \\
\midrule

\multicolumn{5}{l}{\textbf{Global VLM}} \\
Merlin \cite{merlin} & $0.7729{\scriptstyle \pm 0.0062}$ & - & - & $0.6919{\scriptstyle \pm 0.0058}$ \\
CT-CLIP \cite{ct_clip} & - & $0.7082{\scriptstyle \pm 0.0037}$ & $0.5822{\scriptstyle \pm 0.0121}$ & - \\
\midrule
\multicolumn{5}{l}{\textbf{Global Adaptation}} \\
Spatial Transformer & $0.7840{\scriptstyle \pm 0.0055}$ & $0.6467{\scriptstyle \pm 0.0043}$ & $0.5605{\scriptstyle \pm 0.0126}$ & $0.7003{\scriptstyle \pm 0.0056}$ & 0.6729 \\
\midrule
\multicolumn{5}{l}{\textbf{Fine-grained Adaptation}} \\
MLP & $\underline{0.7981}{\scriptstyle \pm 0.0057}$ & $0.6129{\scriptstyle \pm 0.0041}$ & $0.5699{\scriptstyle \pm 0.0121}$ & $0.7389{\scriptstyle \pm 0.0048}$ & 0.6800\\
fVLM$^\dagger$ \cite{fvlm} & $0.7699{\scriptstyle \pm 0.0058}$ & $\underline{0.6842}{\scriptstyle \pm 0.0040}$ & $0.5587{\scriptstyle \pm 0.0124}$ & $0.7036{\scriptstyle \pm 0.0049}$ & 0.6791\\
ViSD-Boost$^\dagger$ \cite{visd_boost} & $0.7803{\scriptstyle \pm 0.0059}$ & $0.6511{\scriptstyle \pm 0.0042}$ & $\underline{0.6139}{\scriptstyle \pm 0.0115}$ & $\mathbf{0.7413}{\scriptstyle \pm 0.0048}$ & \underline{0.6967}\\

\midrule
\multicolumn{5}{l}{\textbf{Global + Fine-grained Adaptation}} \\
ACA &
$\mathbf{0.8213}{\scriptstyle \pm 0.0052}$ &
$\mathbf{0.7311}{\scriptstyle \pm 0.0036}$ &
$\mathbf{0.6723}{\scriptstyle \pm 0.0117}$ &
$\underline{0.7400}{\scriptstyle \pm 0.0049}$ & \textbf{0.7412}\\

\hline
\end{tabular}
}
\end{table}

\section{Results}


\begin{table}[t]
\centering
\caption{Ablation results for ACA across the Merlin and CT-RATE datasets, comparing loss components ($\mathcal{L}_{\text{anatomy}}$, $\mathcal{L}_{\text{scan}}$) and anatomy-guided inference-time pooling (Section~\ref{supp:anatomy_guided}). Merlin* and CT-RATE* denote out-of-distribution evaluation.}
\label{tab:auroc_ablation_comparison}
\resizebox{\textwidth}{!}{%
\begin{tabular}{lcc|cc|c}
\hline
& \multicolumn{2}{c|}{In-distribution} &
\multicolumn{2}{c}{Out-of-distribution} \\
Model & Merlin & CT-RATE & Merlin* & CT-RATE* & Average \\
\midrule

ACA w/o $\mathcal{L}_{\text{anatomy}}$ & $0.7941{\scriptstyle \pm 0.0057}$ & $0.7117{\scriptstyle \pm 0.0037}$ & $0.6130{\scriptstyle \pm 0.0116}$ & $0.7143{\scriptstyle \pm 0.0059}$ & 0.7083\\

ACA w/o $\mathcal{L}_{\text{scan}}$ & $0.8108{\scriptstyle \pm 0.0053}$ & $0.6935{\scriptstyle \pm 0.0040}$ & $0.6364{\scriptstyle \pm 0.0120}$ & $0.7068{\scriptstyle \pm 0.0045}$ & 0.7119\\

\hspace{1em} \textit{+ Anatomy-Guided} &
$\underline{0.8222}{\scriptstyle \pm 0.0052}$ &
$\underline{0.7365}{\scriptstyle \pm 0.0036}$ &
$0.6399{\scriptstyle \pm 0.0120}$ &
$\underline{0.7451}{\scriptstyle \pm 0.0045}$ & 0.7359\\

ACA &
$0.8213{\scriptstyle \pm 0.0052}$ &
$0.7311{\scriptstyle \pm 0.0036}$ &
$\mathbf{0.6723}{\scriptstyle \pm 0.0117}$ &
$0.7400{\scriptstyle \pm 0.0049}$ & \underline{0.7412}\\

\hspace{1em} \textit{+ Anatomy-Guided} &
$\mathbf{0.8372}{\scriptstyle \pm 0.0048}$ &
$\mathbf{0.7413}{\scriptstyle \pm 0.0035}$ &
$\underline{0.6718}{\scriptstyle \pm 0.0118}$ &
$\mathbf{0.7530}{\scriptstyle \pm 0.0048}$ & $\mathbf{0.7508}$\\

\hline
\end{tabular}
}
\end{table}

We evaluate ACA on two CT datasets, Merlin and CT-RATE, using their respective frozen foundation models, Merlin and CT-CLIP, as backbones. For ACA and each baseline (Section~\ref{sec:baselines}), we assess zero-shot classification performance in both in-distribution (i.e., train/test on same dataset) and out-of-distribution (i.e., train/test on different datasets) settings (Section~\ref{sec:eval}). 
We report AUROC as our evaluation metric, as it is threshold-independent and robust to the class imbalance present across both datasets, where most findings have substantially fewer positive than negative cases (Appendix Tables \ref{tab:finding_counts_ct_rate} and \ref{tab:finding_counts_merlin}). Table \ref{tab:auroc_comparison} reports macro-average AUROC across findings for all models, datasets, and distribution settings. To quantify uncertainty, we estimate the mean and standard deviation of macro-average AUROC over 1000 bootstrap resamples.

\textbf {ACA outperforms both global and fine-grained baselines.}
ACA improves over both global VLM foundation models in every setting, raising in-distribution average AUROC from 0.7729 to 0.8213 on Merlin and from 0.7082 to 0.7311 on CT-RATE, with larger gains out-of-distribution (0.5822 to 0.6723 on Merlin*, 0.6919 to 0.7400 on CT-RATE*; note that out-of-distribution evaluation is restricted to the 7 shared findings, see also Appendix Tables \ref{tab:merlin_id}, \ref{tab:ctrate_id}). ACA likewise outperforms the fine-grained baselines, with an average boost in AUROC across settings of 0.061, 0.062, and 0.044 compared to MLP, fVLM, and ViSD-Boost, respectively. Gains are observed up to 11.8 points in-distribution (CT-RATE, vs.\ MLP) and 11.4 points out-of-distribution (Merlin*, vs.\ fVLM); the one exception is a near-tie with MLP and ViSD-Boost on CT-RATE*, which we revisit below. Furthermore, ACA outperforms the global adaptation strategy (Spatial Transformer) in each setting, with an average increase of 0.068 AUROC.

Figure~\ref{fig:per_finding_auroc} illustrates the performance changes per finding compared to the original global VLM models, with finding-level results for all models contained in Appendix Tables \ref{tab:merlin_id}, \ref{tab:ctrate_id}, \ref{tab:merlin_ood}, and \ref{tab:ctrate_ood}. The boosts compared to Merlin and CT-RATE highlight the benefits of structured, anatomy-level modeling, where the findings with the largest performance boosts are often subtle, single organ pathologies (e.g., aortic valve calcification, renal cyst, hepatomegaly), that might get washed out in global, scan-level embedding. Conversely, more diffuse or non-anatomy specific findings show less benefits (e.g., free air, submucosal edema, thrombosis). Compared to fine-grained adaptation alone, the benefits of the global inter-anatomy transformer are apparent in findings such as arterial wall calcification (e.g., 0.13 average AUROC boost (Table \ref{tab:ctrate_id})), that can occur in any artery and thus requires integrating features throughout the scan, as well as organomegaly findings.

\textbf {Ablation of combined loss function.}
Along with the MLP and Spatial Transformer baselines, which ablate ACA's global and anatomy-level architectural components respectively, we performed ablations of ACA's loss function itself (Table \ref{tab:auroc_ablation_comparison}). Removing either the anatomy-level loss (ACA w/o $\mathcal{L}_{\text{anatomy}}$) or the scan-level loss (ACA w/o $\mathcal{L}_{\text{scan}}$) decreases performance (-0.033 and -0.029 average AUROC, respectively), indicating that both anatomy-level and scan-level supervision is important even when using the ACA architecture fixed.

\begin{figure}[t]
    \centering
    \includegraphics[width=\linewidth]{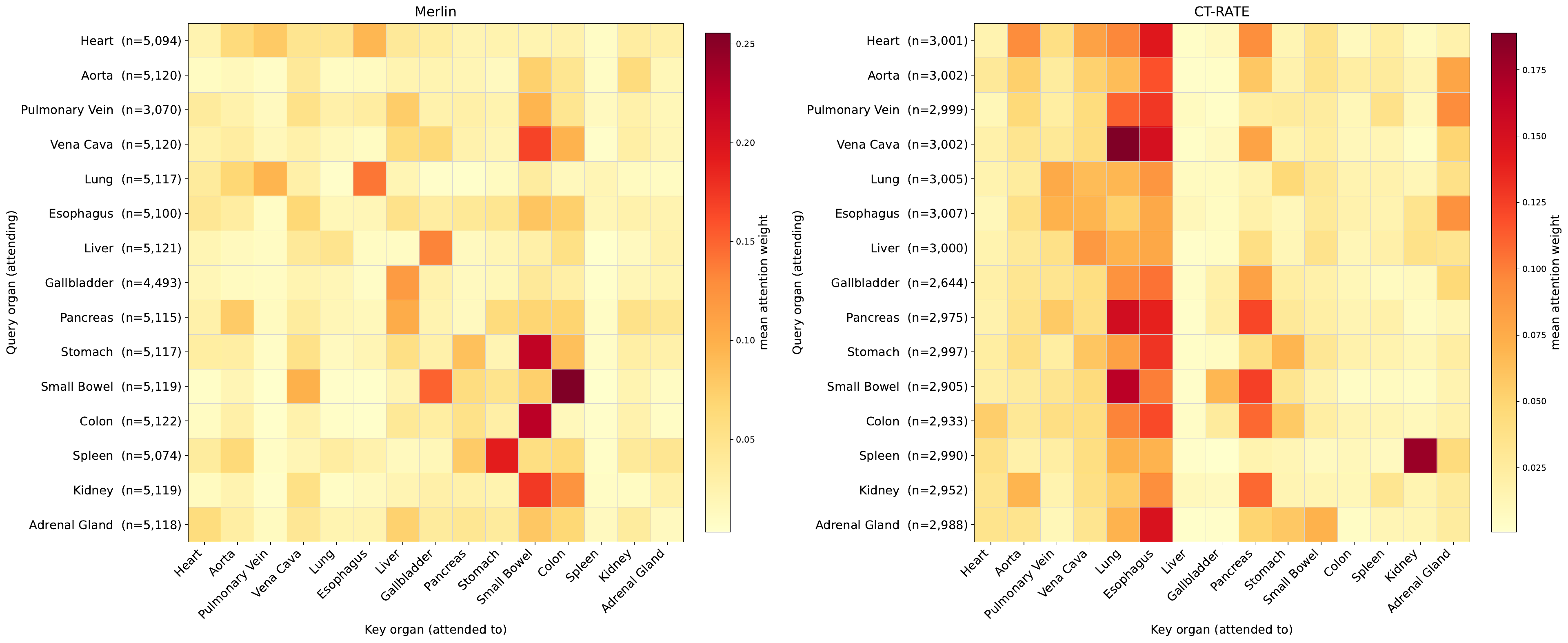}
    \caption{
    Each heatmap shows the mean attention weights between fifteen major anatomical structures, averaged over all transformer layers, attention heads, and test scans. Left: ACA model trained on Merlin and evaluated on the Merlin test set. Right: ACA model trained on CT-RATE and evaluated on the CT-RATE test set. Sample counts per query anatomy reflect each respective dataset. 
    }
    \label{fig:attention}
\end{figure}

\textbf {Anatomy-guided pooling.}
As the results thus far have used ACA's scan-level representation, consisting of mean pooling over the contextualized anatomy embeddings, we also explored an inference-time strategy that assigns different pooling weights depending on the tested finding (see Appendix \ref{supp:anatomy_guided}). Upweighting the pooling of anatomical structures relevant to each finding (Table \ref{tab:auroc_ablation_comparison}) further improves AUROC for both ACA and ACA w/o $\mathcal{L}_{\text{scan}}$ in most settings. For ACA, anatomy-guided pooling raises in-distribution AUROC from 0.8213 to 0.8372 on Merlin and from 0.7311 to 0.7413 on CT-RATE, and out-of-distribution AUROC from 0.7400 to 0.7530 on CT-RATE*, while leaving Merlin* essentially unchanged (0.6723 vs.\ 0.6718). For ACA w/o $\mathcal{L}_{\text{scan}}$, the effect is larger, in-distribution AUROC improves from 0.6935 to 0.7365 on CT-RATE, and out-of-distribution AUROC improves on both Merlin* (0.6364 vs.\ 0.6399) and CT-RATE* (0.7068 vs.\ 0.7451). These results suggest that restricting attention to clinically relevant anatomy is most useful when the model lacks scan-level supervision to otherwise aggregate global context.

\begin{figure}[t]
    \centering
    \includegraphics[width=0.9\linewidth]{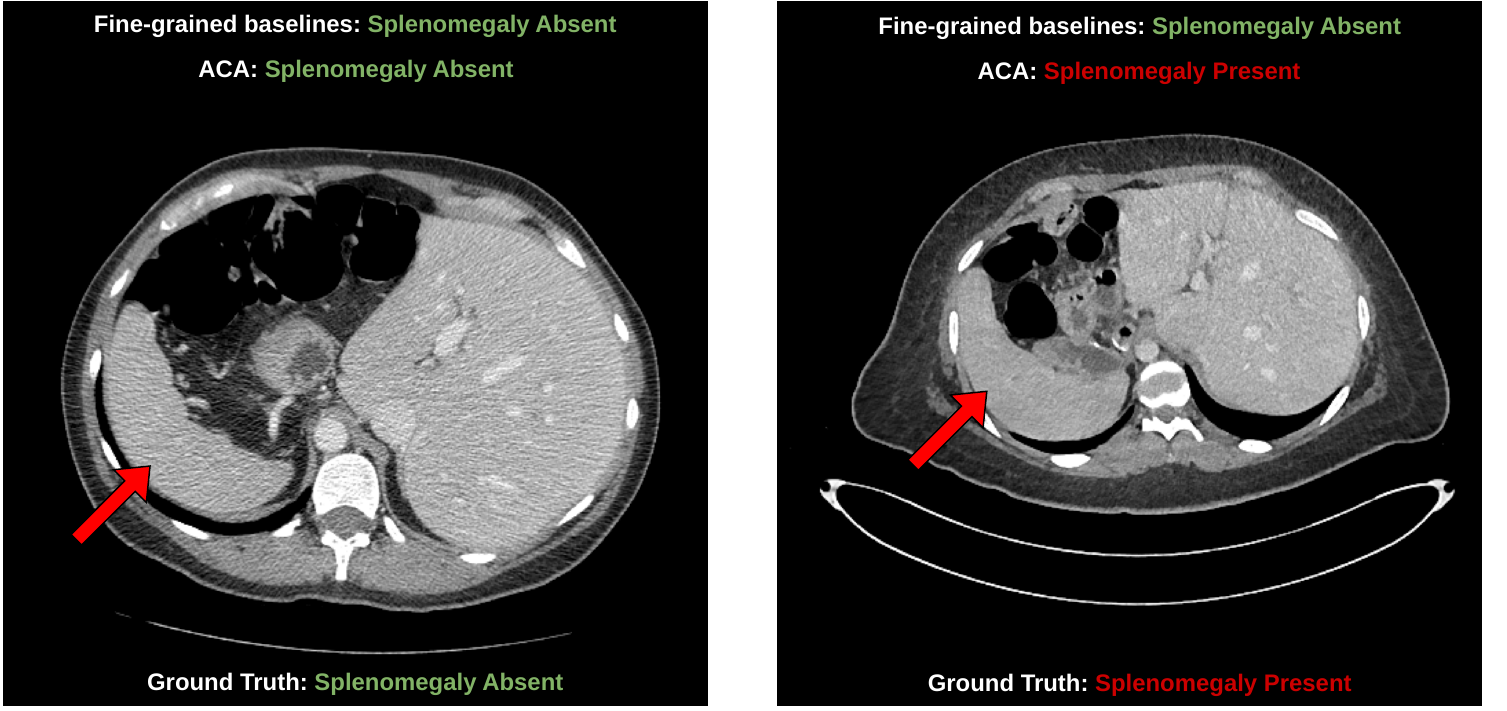}
    \caption{Two CT scans from the Merlin dataset. The right patient has splenomegaly and the left does not. Despite the spleens appearing of comparable size in isolation, ACA correctly identifies the right patient as positive while all fine-grained baselines predict negative. The relative proportions of surrounding organs provide a discriminative signal that single-organ embeddings cannot capture.
    }
    \label{fig:context}
\end{figure}

\textbf {Learned Anatomy Associations.}
Using the inter-anatomy transformer, we visualize the attention to understand any inter-anatomy relationships the model implicitly learns.  Figure~\ref{fig:attention} shows the average attention weights between key organs across the test set for ACA models trained on Merlin (left) and CT-RATE (right). We find that both models learn anatomically plausible cross-anatomy associations that emerge purely from contrastive training, and the specific associations each model learns track the anatomical scope of its training dataset. Merlin is an abdominal CT cohort with findings concentrated in abdominal and GI pathology (Appendix Table~\ref{tab:finding_counts_merlin}), which is reflected in its strongest learned associations. The largest attention weights are amongst the stomach, small bowel, and colon, reflecting their continuity along the GI tract. Additionally, the liver and gallbladder attend to one another with the highest weight in each anatomy's row, mirroring their biliary connection. CT-RATE, by contrast, is a chest CT dataset with findings concentrated in cardiopulmonary pathology (Appendix Table~\ref{tab:finding_counts_ct_rate}), which is reflected in the strong attention weights directed to the lung and esophagus in the CT-RATE ACA model. 
The spleen also attends most strongly to the kidney and the kidney to the pancreas, structures that sit directly adjacent to one another behind the abdominal cavity.

Another notable observation is that self-attention along the diagonal is relatively weak in both datasets. We hypothesize that the fine-grained adaptation in ACA, built upon frozen foundation models, already provides strong anatomy-specific representations, reducing the need for the transformer to reinforce anatomy identity through self-attention. Instead, the adaptation module appears to focus on modeling inter-anatomy relationships, consistent with prior work viewing attention as a mechanism for information routing and interaction modeling rather than self-reinforcement \cite{abnar2020quantifying}.

\textbf {Contextual Anatomy Reasoning.}
Many findings in the dataset require cross-anatomy context to detect reliably. 
A clear case study is organomegaly, where the challenge is not detecting a structure's presence but judging its relative size. For instance, a spleen of a given absolute cross-sectional area may be pathological in one patient and entirely normal in another, depending on body habitus and the proportions of surrounding structures. Figure \ref{fig:context} illustrates this, where two scans in the Merlin test set contain spleens of similar cross-section area, yet only one carries a splenomegaly label. Baselines that embed each organ independently have no mechanism to make this comparison, as each anatomy's representation is formed without reference to neighboring structures. ACA's inter-anatomy transformer, by attending jointly over all organ tokens in the same forward pass, can represent the relative size of each structure with respect to the others. The contrastive objective then ties this relational representation to anatomy-level text that naturally expresses such comparisons (\emph{``the spleen is enlarged''}), grounding the model in contextual reasoning and facilitating a correct splenomegaly prediction for the example on the right at inference.

\section{Discussion}

ACA adapts frozen CT foundation model representations to provide structured, anatomy-level embeddings while facilitating scan-level, contextual reasoning with a lightweight inter-anatomy transformer. ACA consistently improves zero-shot finding classification over global vision-language models and existing fine-grained adaptation methods, in-distribution and out-of-distribution across two large-scale datasets. Ablations confirm that both core components, anatomical decomposition and scan-level report supervision, contribute independently to this improvement, and that restricting pooling to clinically relevant anatomical structures at inference time can provide an additional low-cost gain. The attention weights learned by the inter-anatomy transformer also reflect plausible associations and underlying dataset characteristics.


\textbf {Computational Benefits of Adaptation.}
ACA is designed to adapt frozen foundation model representations with lightweight trainable modules, avoiding the computational cost of full end-to-end retraining that has traditionally been performed for fine-grained modeling \cite{fvlm, visd_boost, ct_glip}. A natural alternative is to fine-tune the backbone directly, which would significantly increase compute but could further improve performance. To explore this potential, we developed a LoRA-finetuned~\cite{hu2022lora} variant of the full Merlin-based ACA model, trained end-to-end using the same anatomy decomposition and contrastive objective described in Section~\ref{sec:loss}. The LoRA variant achieved an increase in average AUROC of only 0.008 on the Merlin test set compared to the original ACA formulation with the same hyperparameters. This suggests that the lightweight ACA adaptation of frozen foundation model representations recovers the majority of the benefit of full end-to-end training at a fraction of the computational cost, consistent with a broader trend in parameter-efficient  adaptation \cite{chen2022adaptformer, he2023parameter, liu2022polyhistor}. A further practical advantage of this design is its modularity: because the adaptation module is decoupled from the backbone, it can be applied to any CT foundation model at low additional cost.

\textbf {Limitations.}
Our comparisons to fVLM \cite{fvlm} and ViSD-Boost \cite{visd_boost} are reimplementations adapted to operate on frozen foundation model embeddings, rather than the original end-to-end trained methods. This isolates each method's alignment strategy under a matched, frozen-embedding setting, but means our results support a narrower claim than outperforming fVLM and ViSD-Boost as originally published. A full end-to-end retraining of these methods would be needed to compare against their originally reported performance. Additionally, ACA's anatomy decomposition depends on TotalSegmentator's vocabulary, so structures outside it are not directly represented. Per-anatomy findings and normality labels used during training are extracted automatically with an LLM rather than verified by radiologists, which may introduce label noise, though all evaluations use each dataset's ground-truth scan-level labels. Our evaluation is limited to two datasets and backbones, with out-of-distribution comparisons restricted to the 7 findings shared between Merlin and CT-RATE, so broader generalization remains untested. Finally, this work addresses only zero-shot finding classification, and extending ACA to tasks such as segmentation, outcome prediction, and report generation is left to future work.

\textbf {Conclusion.}
ACA combines fine-grained alignment with global contextualization for CT vision-language modeling by adapting existing foundation models rather than training from scratch. 
We see this as a practical path toward improving representation learning in a compute-efficient manner. 

\section*{Acknowledgements}
W.L. gratefully acknowledges funding support from the Ellison Foundation.

%
%
\bibliographystyle{splncs04}
\bibliography{main}

@String(ICLR  = {Int. Conf. Learn. Represent.})

@String(AAAI  = {AAAI})

@String(ICLR  = {ICLR})

@article{merlin,
  title={Merlin: a computed tomography vision--language foundation model and dataset},
  author={Blankemeier, Louis and Kumar, Ashwin and Cohen, Joseph Paul and Liu, Jiaming and Liu, Longchao and Van Veen, Dave and Gardezi, Syed Jamal Safdar and Yu, Hongkun and Paschali, Magdalini and Chen, Zhihong and others},
  journal={Nature},
  pages={1--11},
  year={2026},
  publisher={Nature Publishing Group UK London}
}

@article{ct_clip,
  title={Generalist foundation models from a multimodal dataset for 3D computed tomography},
  author={Hamamci, Ibrahim Ethem and Er, Sezgin and Wang, Chenyu and Almas, Furkan and Simsek, Ayse Gulnihan and Esirgun, Sevval Nil and Dogan, Irem and Durugol, Omer Faruk and Hou, Benjamin and Shit, Suprosanna and others},
  journal={Nature Biomedical Engineering},
  pages={1--19},
  year={2026},
  publisher={Nature Publishing Group UK London}
}

@article{fvlm,
  title={Large-scale and fine-grained vision-language pre-training for enhanced ct image understanding},
  author={Shui, Zhongyi and Zhang, Jianpeng and Cao, Weiwei and Wang, Sinuo and Guo, Ruizhe and Lu, Le and Yang, Lin and Ye, Xianghua and Liang, Tingbo and Zhang, Qi and others},
  journal={arXiv preprint arXiv:2501.14548},
  year={2025}
}

@inproceedings{visd_boost,
  title={Boosting vision semantic density with anatomy normality modeling for medical vision-language pre-training},
  author={Cao, Weiwei and Zhang, Jianpeng and Shui, Zhongyi and Wang, Sinuo and Chen, Zeli and Li, Xi and Lu, Le and Ye, Xianghua and Zhang, Qi and Liang, Tingbo and others},
  booktitle={Proceedings of the IEEE/CVF International Conference on Computer Vision},
  pages={23041--23050},
  year={2025}
}

@article{ct_glip,
  title={Ct-glip: 3d grounded language-image pretraining with ct scans and radiology reports for full-body scenarios},
  author={Lin, Jingyang and Xia, Yingda and Zhang, Jianpeng and Yan, Ke and Cao, Kai and Lu, Le and Luo, Jiebo and Zhang, Ling},
  journal={arXiv preprint arXiv:2404.15272},
  year={2024}
}

@article{percival ,
  title={A Pan-Organ Vision-Language Model for Generalizable 3D CT Representations},
  author={Beeche, Cameron and Kim, Joonghyun and Tavolinejad, Hamed and Zhao, Bingxin and Sharma, Rakesh and Duda, Jeffrey and Gee, James and Dako, Farouk and Verma, Anurag and Morse, Colleen and others},
  journal={medRxiv},
  year={2025}
}

@article{wasserthal2023totalsegmentator,
  title={TotalSegmentator: robust segmentation of 104 anatomic structures in CT images},
  author={Wasserthal, Jakob and Breit, Hanns-Christian and Meyer, Manfred T and Pradella, Maurice and Hinck, Daniel and Sauter, Alexander W and Heye, Tobias and Boll, Daniel T and Cyriac, Joshy and Yang, Shan and others},
  journal={Radiology: Artificial Intelligence},
  volume={5},
  number={5},
  pages={e230024},
  year={2023},
  publisher={Radiological Society of North America}
}

@article{yang2025qwen3,
  title={Qwen3 technical report},
  author={Yang, An and Li, Anfeng and Yang, Baosong and Zhang, Beichen and Hui, Binyuan and Zheng, Bo and Yu, Bowen and Gao, Chang and Huang, Chengen and Lv, Chenxu and others},
  journal={arXiv preprint arXiv:2505.09388},
  year={2025}
}

@article{3dino,
  title={A generalizable 3D framework and model for self-supervised learning in medical imaging},
  author={Xu, Tony and Hosseini, Sepehr and Anderson, Chris and Rinaldi, Anthony and Krishnan, Rahul G and Martel, Anne L and Goubran, Maged},
  journal={npj Digital Medicine},
  volume={8},
  number={1},
  pages={639},
  year={2025},
  publisher={Nature Publishing Group UK London}
}

@article{ct_fm,
  title={Vision foundation models for computed tomography},
  author={Pai, Suraj and Hadzic, Ibrahim and Bontempi, Dennis and Bressem, Keno and Kann, Benjamin H and Fedorov, Andriy and Mak, Raymond H and Aerts, Hugo JWL},
  journal={arXiv preprint arXiv:2501.09001},
  year={2025}
}

@article{lct_found,
  title={A lung CT vision foundation model facilitating disease diagnosis and medical imaging},
  author={Gao, Zebin and Zhang, Guoxun and Liang, Hengrui and Liu, Jiaxin and Ma, Liangdi and Wang, Tianyun and Guo, Yanchen and Chen, YuJia and Yan, Zeping and Chen, Xiangru and others},
  journal={Nature Communications},
  year={2025},
  publisher={Nature Publishing Group UK London}
}

@article{m3fm,
  title={Medical multimodal multitask foundation model for lung cancer screening},
  author={Niu, Chuang and Lyu, Qing and Carothers, Christopher D and Kaviani, Parisa and Tan, Josh and Yan, Pingkun and Kalra, Mannudeep K and Whitlow, Christopher T and Wang, Ge},
  journal={Nature Communications},
  volume={16},
  number={1},
  pages={1523},
  year={2025},
  publisher={Nature Publishing Group UK London}
}

@inproceedings{vista3d,
  title={VISTA3D: A unified segmentation foundation model for 3D medical imaging},
  author={He, Yufan and Guo, Pengfei and Tang, Yucheng and Myronenko, Andriy and Nath, Vishwesh and Xu, Ziyue and Yang, Dong and Zhao, Can and Simon, Benjamin and Belue, Mason and others},
  booktitle={Proceedings of the Computer Vision and Pattern Recognition Conference},
  pages={20863--20873},
  year={2025}
}

@inproceedings{ct_vit,
  title={Generatect: Text-conditional generation of 3d chest ct volumes},
  author={Hamamci, Ibrahim Ethem and Er, Sezgin and Sekuboyina, Anjany and Simsar, Enis and Tezcan, Alperen and Simsek, Ayse Gulnihan and Esirgun, Sevval Nil and Almas, Furkan and Do{\u{g}}an, Irem and Dasdelen, Muhammed Furkan and others},
  booktitle={European Conference on Computer Vision},
  pages={126--143},
  year={2024},
  organization={Springer}
}

@article{hu2022lora,
  title={Lora: Low-rank adaptation of large language models.},
  author={Hu, Edward J and Shen, Yelong and Wallis, Phillip and Allen-Zhu, Zeyuan and Li, Yuanzhi and Wang, Shean and Wang, Liang and Chen, Weizhu and others},
  journal={Iclr},
  volume={1},
  number={2},
  pages={3},
  year={2022}
}

@article{chen2022adaptformer,
  title={Adaptformer: Adapting vision transformers for scalable visual recognition},
  author={Chen, Shoufa and Ge, Chongjian and Tong, Zhan and Wang, Jiangliu and Song, Yibing and Wang, Jue and Luo, Ping},
  journal={Advances in Neural Information Processing Systems},
  volume={35},
  pages={16664--16678},
  year={2022}
}

@inproceedings{cao2024bootstrapping,
  title={Bootstrapping chest ct image understanding by distilling knowledge from x-ray expert models},
  author={Cao, Weiwei and Zhang, Jianpeng and Xia, Yingda and Mok, Tony CW and Li, Zi and Ye, Xianghua and Lu, Le and Zheng, Jian and Tang, Yuxing and Zhang, Ling},
  booktitle={Proceedings of the IEEE/CVF Conference on Computer Vision and Pattern Recognition},
  pages={11238--11247},
  year={2024}
}

@inproceedings{huang2021gloria,
  title={Gloria: A multimodal global-local representation learning framework for label-efficient medical image recognition},
  author={Huang, Shih-Cheng and Shen, Liyue and Lungren, Matthew P and Yeung, Serena},
  booktitle={Proceedings of the IEEE/CVF international conference on computer vision},
  pages={3942--3951},
  year={2021}
}

@inproceedings{muller2022joint,
  title={Joint learning of localized representations from medical images and reports},
  author={M{\"u}ller, Philip and Kaissis, Georgios and Zou, Congyu and Rueckert, Daniel},
  booktitle={European conference on computer vision},
  pages={685--701},
  year={2022},
  organization={Springer}
}

@article{wang2022multi,
  title={Multi-granularity cross-modal alignment for generalized medical visual representation learning},
  author={Wang, Fuying and Zhou, Yuyin and Wang, Shujun and Vardhanabhuti, Varut and Yu, Lequan},
  journal={Advances in neural information processing systems},
  volume={35},
  pages={33536--33549},
  year={2022}
}

@inproceedings{he2023parameter,
  title={Parameter-efficient model adaptation for vision transformers},
  author={He, Xuehai and Li, Chunyuan and Zhang, Pengchuan and Yang, Jianwei and Wang, Xin Eric},
  booktitle={Proceedings of the AAAI Conference on Artificial Intelligence},
  volume={37},
  number={1},
  pages={817--825},
  year={2023}
}

@article{liu2022polyhistor,
  title={Polyhistor: Parameter-efficient multi-task adaptation for dense vision tasks},
  author={Liu, Yen-Cheng and Ma, Chih-Yao and Tian, Junjiao and He, Zijian and Kira, Zsolt},
  journal={Advances in neural information processing systems},
  volume={35},
  pages={36889--36901},
  year={2022}
}

@inproceedings{abnar2020quantifying,
  title={Quantifying attention flow in transformers},
  author={Abnar, Samira and Zuidema, Willem},
  booktitle={Proceedings of the 58th annual meeting of the association for computational linguistics},
  pages={4190--4197},
  year={2020}
}

@article{zhou2021models,
  title={Models genesis},
  author={Zhou, Zongwei and Sodha, Vatsal and Pang, Jiaxuan and Gotway, Michael B and Liang, Jianming},
  journal={Medical image analysis},
  volume={67},
  pages={101840},
  year={2021},
  publisher={Elsevier}
}

@inproceedings{xie2022unimiss,
  title={Unimiss: Universal medical self-supervised learning via breaking dimensionality barrier},
  author={Xie, Yutong and Zhang, Jianpeng and Xia, Yong and Wu, Qi},
  booktitle={European Conference on Computer Vision},
  pages={558--575},
  year={2022},
  organization={Springer}
}

@article{wu2025towards,
  title={Towards generalist foundation model for radiology by leveraging web-scale 2d\&3d medical data},
  author={Wu, Chaoyi and Zhang, Xiaoman and Zhang, Ya and Hui, Hui and Wang, Yanfeng and Xie, Weidi},
  journal={Nature Communications},
  volume={16},
  number={1},
  pages={7866},
  year={2025},
  publisher={Nature Publishing Group UK London}
}

@inproceedings{wu2024voco,
  title={Voco: A simple-yet-effective volume contrastive learning framework for 3d medical image analysis},
  author={Wu, Linshan and Zhuang, Jiaxin and Chen, Hao},
  booktitle={Proceedings of the IEEE/CVF conference on computer vision and pattern recognition},
  pages={22873--22882},
  year={2024}
}

@article{wang2025sam,
  title={SAM-Med3D: a vision foundation model for general-purpose segmentation on volumetric medical images},
  author={Wang, Haoyu and Guo, Sizheng and Ye, Jin and Deng, Zhongying and Cheng, Junlong and Li, Tianbin and Chen, Jianpin and Su, Yanzhou and Huang, Ziyan and Shen, Yiqing and others},
  journal={IEEE Transactions on Neural Networks and Learning Systems},
  year={2025},
  publisher={IEEE}
}

@article{zhang2026gcl,
  title={CA-GCL: Cross-Anatomy Global-Local Contrastive Learning for Robust 3D Medical Image Understanding},
  author={Zhang, Hanwen and Liu, Yao and Dai, Die and Yang, Jiaye and Liu, Qiao and Xie, Yutong and Wang, Peng},
  journal={arXiv preprint arXiv:2605.13544},
  year={2026}
}

@article{wen2024biological,
  title={Biological age shows that no organ system is an island},
  author={Wen, Junhao},
  journal={Nature},
  volume={4},
  pages={1182--1183},
  year={2024}
}

@inproceedings{radford2021learning,
  title={Learning transferable visual models from natural language supervision},
  author={Radford, Alec and Kim, Jong Wook and Hallacy, Chris and Ramesh, Aditya and Goh, Gabriel and Agarwal, Sandhini and Sastry, Girish and Askell, Amanda and Mishkin, Pamela and Clark, Jack and others},
  booktitle={International conference on machine learning},
  pages={8748--8763},
  year={2021},
  organization={PmLR}
}

@article{bao2021beit,
  title={Beit: Bert pre-training of image transformers},
  author={Bao, Hangbo and Dong, Li and Piao, Songhao and Wei, Furu},
  journal={arXiv preprint arXiv:2106.08254},
  year={2021}
}

@inproceedings{he2020momentum,
  title={Momentum contrast for unsupervised visual representation learning},
  author={He, Kaiming and Fan, Haoqi and Wu, Yuxin and Xie, Saining and Girshick, Ross},
  booktitle={Proceedings of the IEEE/CVF conference on computer vision and pattern recognition},
  pages={9729--9738},
  year={2020}
}

@inproceedings{caron2021emerging,
  title={Emerging properties in self-supervised vision transformers},
  author={Caron, Mathilde and Touvron, Hugo and Misra, Ishan and J{\'e}gou, Herv{\'e} and Mairal, Julien and Bojanowski, Piotr and Joulin, Armand},
  booktitle={Proceedings of the IEEE/CVF international conference on computer vision},
  pages={9650--9660},
  year={2021}
}

@inproceedings{he2022masked,
  title={Masked autoencoders are scalable vision learners},
  author={He, Kaiming and Chen, Xinlei and Xie, Saining and Li, Yanghao and Doll{\'a}r, Piotr and Girshick, Ross},
  booktitle={Proceedings of the IEEE/CVF conference on computer vision and pattern recognition},
  pages={16000--16009},
  year={2022}
}

@inproceedings{xie2022simmim,
  title={Simmim: A simple framework for masked image modeling},
  author={Xie, Zhenda and Zhang, Zheng and Cao, Yue and Lin, Yutong and Bao, Jianmin and Yao, Zhuliang and Dai, Qi and Hu, Han},
  booktitle={Proceedings of the IEEE/CVF conference on computer vision and pattern recognition},
  pages={9653--9663},
  year={2022}
}

@inproceedings{chen2020simple,
  title={A simple framework for contrastive learning of visual representations},
  author={Chen, Ting and Kornblith, Simon and Norouzi, Mohammad and Hinton, Geoffrey},
  booktitle={International conference on machine learning},
  pages={1597--1607},
  year={2020},
  organization={PmLR}
}

@inproceedings{wang2022medclip,
  title={Medclip: Contrastive learning from unpaired medical images and text},
  author={Wang, Zifeng and Wu, Zhenbang and Agarwal, Dinesh and Sun, Jimeng},
  booktitle={Proceedings of the 2022 Conference on Empirical Methods in Natural Language Processing},
  pages={3876--3887},
  year={2022}
}

@inproceedings{wu2023medklip,
  title={Medklip: Medical knowledge enhanced language-image pre-training for x-ray diagnosis},
  author={Wu, Chaoyi and Zhang, Xiaoman and Zhang, Ya and Wang, Yanfeng and Xie, Weidi},
  booktitle={Proceedings of the IEEE/CVF international conference on computer vision},
  pages={21372--21383},
  year={2023}
}

@inproceedings{carreira2017quo,
  title={Quo vadis, action recognition? a new model and the kinetics dataset},
  author={Carreira, Joao and Zisserman, Andrew},
  booktitle={proceedings of the IEEE Conference on Computer Vision and Pattern Recognition},
  pages={6299--6308},
  year={2017}
}

\clearpage
\appendix
\section{Preprocessing}

\small

\begin{longtable}{r p{4.5cm} p{4.5cm} r}
\caption{Mapping of TotalSegmentator labels to grouped anatomical structures for fine-grained modeling.}\\
\toprule
\textbf{Idx} & \textbf{TotalSegmentator Name} & \textbf{Grouping Name} & \textbf{Grp} \\
\midrule
\endfirsthead

\toprule
\textbf{Idx} & \textbf{TotalSegmentator Name} & \textbf{Grouping Name} & \textbf{Grp} \\
\midrule
\endhead

\bottomrule
\endfoot

1 & spleen & spleen & 1 \\
2 & kidney\_right & kidney & 2 \\
3 & kidney\_left & kidney & 2 \\
4 & gallbladder & gallbladder & 3 \\
5 & liver & liver & 4 \\
6 & stomach & stomach & 5 \\
7 & pancreas & pancreas & 6 \\
8 & adrenal\_gland\_right & adrenal\_gland & 7 \\
9 & adrenal\_gland\_left & adrenal\_gland & 7 \\
10 & lung\_upper\_lobe\_left & lung & 8 \\
11 & lung\_lower\_lobe\_left & lung & 8 \\
12 & lung\_upper\_lobe\_right & lung & 8 \\
13 & lung\_middle\_lobe\_right & lung & 8 \\
14 & lung\_lower\_lobe\_right & lung & 8 \\
15 & esophagus & esophagus & 9 \\
16 & trachea & trachea & 10 \\
17 & thyroid\_gland & thyroid\_gland & 11 \\
18 & small\_bowel & small\_bowel & 12 \\
19 & duodenum & small\_bowel & 12 \\
20 & colon & colon & 13 \\
21 & urinary\_bladder & urinary\_bladder & 14 \\
22 & prostate & prostate & 15 \\
23 & kidney\_cyst\_left & kidney & 2 \\
24 & kidney\_cyst\_right & kidney & 2 \\
25 & sacrum & sacrum & 16 \\
26 & vertebrae\_S1 & sacrum & 16 \\
27 & vertebrae\_L5 & lumbar\_vertebrae & 17 \\
28 & vertebrae\_L4 & lumbar\_vertebrae & 17 \\
29 & vertebrae\_L3 & lumbar\_vertebrae & 17 \\
30 & vertebrae\_L2 & lumbar\_vertebrae & 17 \\
31 & vertebrae\_L1 & lumbar\_vertebrae & 17 \\
32 & vertebrae\_T12 & thoracic\_vertebrae & 18 \\
33 & vertebrae\_T11 & thoracic\_vertebrae & 18 \\
34 & vertebrae\_T10 & thoracic\_vertebrae & 18 \\
35 & vertebrae\_T9 & thoracic\_vertebrae & 18 \\
36 & vertebrae\_T8 & thoracic\_vertebrae & 18 \\
37 & vertebrae\_T7 & thoracic\_vertebrae & 18 \\
38 & vertebrae\_T6 & thoracic\_vertebrae & 18 \\
39 & vertebrae\_T5 & thoracic\_vertebrae & 18 \\
40 & vertebrae\_T4 & thoracic\_vertebrae & 18 \\
41 & vertebrae\_T3 & thoracic\_vertebrae & 18 \\
42 & vertebrae\_T2 & thoracic\_vertebrae & 18 \\
43 & vertebrae\_T1 & thoracic\_vertebrae & 18 \\
44 & vertebrae\_C7 & cervical\_vertebrae & 19 \\
45 & vertebrae\_C6 & cervical\_vertebrae & 19 \\
46 & vertebrae\_C5 & cervical\_vertebrae & 19 \\
47 & vertebrae\_C4 & cervical\_vertebrae & 19 \\
48 & vertebrae\_C3 & cervical\_vertebrae & 19 \\
49 & vertebrae\_C2 & cervical\_vertebrae & 19 \\
50 & vertebrae\_C1 & cervical\_vertebrae & 19 \\
51 & heart & heart & 20 \\
52 & aorta & aorta & 21 \\
53 & pulmonary\_vein & pulmonary\_vein & 22 \\
54 & brachiocephalic\_trunk & brachiocephalic\_trunk & 23 \\
55 & subclavian\_artery\_right & subclavian\_artery & 24 \\
56 & subclavian\_artery\_left & subclavian\_artery & 24 \\
57 & common\_carotid\_artery\_right & common\_carotid\_artery & 25 \\
58 & common\_carotid\_artery\_left & common\_carotid\_artery & 25 \\
59 & brachiocephalic\_vein\_left & brachiocephalic\_vein & 26 \\
60 & brachiocephalic\_vein\_right & brachiocephalic\_vein & 26 \\
61 & atrial\_appendage\_left & heart & 20 \\
62 & superior\_vena\_cava & vena\_cava & 27 \\
63 & inferior\_vena\_cava & vena\_cava & 27 \\
64 & portal\_vein\_and\_splenic\_vein & portal\_vein\_and\_splenic\_vein & 28 \\
65 & iliac\_artery\_left & iliac\_artery & 29 \\
66 & iliac\_artery\_right & iliac\_artery & 29 \\
67 & iliac\_vena\_left & iliac\_vena & 30 \\
68 & iliac\_vena\_right & iliac\_vena & 30 \\
69 & humerus\_left & humerus & 31 \\
70 & humerus\_right & humerus & 31 \\
71 & scapula\_left & scapula & 32 \\
72 & scapula\_right & scapula & 32 \\
73 & clavicula\_left & clavicula & 33 \\
74 & clavicula\_right & clavicula & 33 \\
75 & femur\_left & femur & 34 \\
76 & femur\_right & femur & 34 \\
77 & hip\_left & hip & 35 \\
78 & hip\_right & hip & 35 \\
79 & spinal\_cord & spinal\_cord & 36 \\
80 & gluteus\_maximus\_left & gluteus & 37 \\
81 & gluteus\_maximus\_right & gluteus & 37 \\
82 & gluteus\_medius\_left & gluteus & 37 \\
83 & gluteus\_medius\_right & gluteus & 37 \\
84 & gluteus\_minimus\_left & gluteus & 37 \\
85 & gluteus\_minimus\_right & gluteus & 37 \\
86 & autochthon\_left & autochthon & 38 \\
87 & autochthon\_right & autochthon & 38 \\
88 & iliopsoas\_left & iliopsoas & 39 \\
89 & iliopsoas\_right & iliopsoas & 39 \\
90 & brain & brain & 40 \\
91 & skull & skull & 41 \\
92 & rib\_left\_1 & rib & 42 \\
93 & rib\_left\_2 & rib & 42 \\
94 & rib\_left\_3 & rib & 42 \\
95 & rib\_left\_4 & rib & 42 \\
96 & rib\_left\_5 & rib & 42 \\
97 & rib\_left\_6 & rib & 42 \\
98 & rib\_left\_7 & rib & 42 \\
99 & rib\_left\_8 & rib & 42 \\
100 & rib\_left\_9 & rib & 42 \\
101 & rib\_left\_10 & rib & 42 \\
102 & rib\_left\_11 & rib & 42 \\
103 & rib\_left\_12 & rib & 42 \\
104 & rib\_right\_1 & rib & 42 \\
105 & rib\_right\_2 & rib & 42 \\
106 & rib\_right\_3 & rib & 42 \\
107 & rib\_right\_4 & rib & 42 \\
108 & rib\_right\_5 & rib & 42 \\
109 & rib\_right\_6 & rib & 42 \\
110 & rib\_right\_7 & rib & 42 \\
111 & rib\_right\_8 & rib & 42 \\
112 & rib\_right\_9 & rib & 42 \\
113 & rib\_right\_10 & rib & 42 \\
114 & rib\_right\_11 & rib & 42 \\
115 & rib\_right\_12 & rib & 42 \\
116 & sternum & sternum & 43 \\
117 & costal\_cartilages & costal\_cartilages & 44 \\

\label{anatomy_mapping}
\end{longtable}

\begin{figure*}[t]
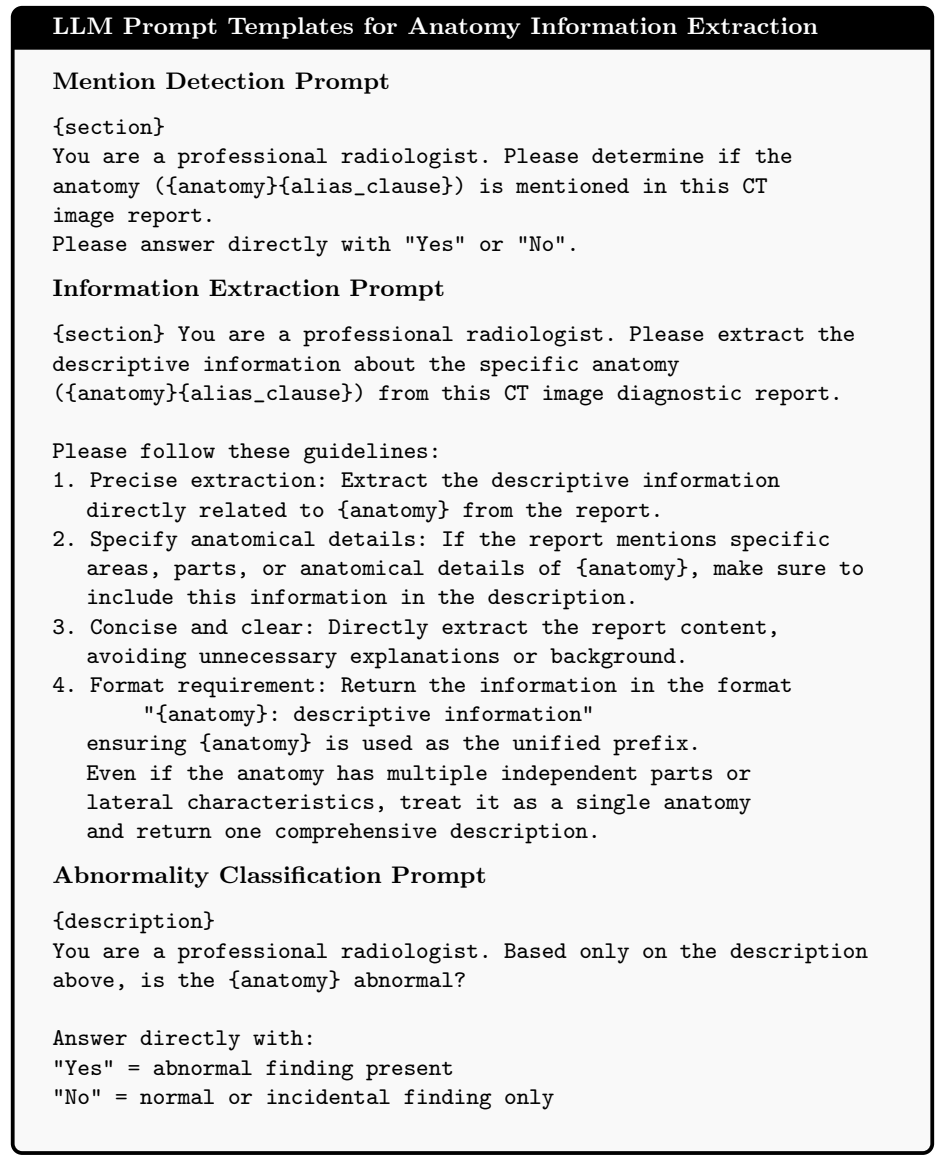

\centering

\begin{tcolorbox}[
    colback=gray!5,
    colframe=black,
    title={LLM Prompt Templates for Anatomy Information Extraction},
    width=\textwidth,
    fonttitle=\bfseries
]

\textbf{Mention Detection Prompt}
\begin{lstlisting}[basicstyle=\ttfamily\small]
{section}
You are a professional radiologist. Please determine if the
anatomy ({anatomy}{alias_clause}) is mentioned in this CT 
image report.
Please answer directly with "Yes" or "No".
\end{lstlisting}

\textbf{Information Extraction Prompt}
\begin{lstlisting}[basicstyle=\ttfamily\small]
{section} You are a professional radiologist. Please extract the 
descriptive information about the specific anatomy 
({anatomy}{alias_clause}) from this CT image diagnostic report.

Please follow these guidelines:
1. Precise extraction: Extract the descriptive information 
   directly related to {anatomy} from the report.
2. Specify anatomical details: If the report mentions specific 
   areas, parts, or anatomical details of {anatomy}, make sure to 
   include this information in the description.
3. Concise and clear: Directly extract the report content, 
   avoiding unnecessary explanations or background.
4. Format requirement: Return the information in the format
        "{anatomy}: descriptive information"
   ensuring {anatomy} is used as the unified prefix.
   Even if the anatomy has multiple independent parts or
   lateral characteristics, treat it as a single anatomy
   and return one comprehensive description.
\end{lstlisting}

\textbf{Abnormality Classification Prompt}
\begin{lstlisting}[basicstyle=\ttfamily\small]
{description}
You are a professional radiologist. Based only on the description
above, is the {anatomy} abnormal?

Answer directly with:
"Yes" = abnormal finding present
"No"  = normal or incidental finding only
\end{lstlisting}

\end{tcolorbox}

\caption{
Prompt templates used for structured anatomy-level information extraction from CT radiology reports. Since reports describe findings globally rather than per organ, a three-stage pipeline is applied. The \textbf{mention detection} prompt screens whether a given anatomy is discussed at all, avoiding spurious extractions for absent organs. The \textbf{information extraction} prompt isolates organ-specific descriptive text, using alias hints and formatting constraints to produce a clean, unified description across sub-structures. The \textbf{abnormality classification} prompt labels the result as normal or abnormal, providing the supervision signal for false negative reduction during training. In all templates, \texttt{\{section\}} is the raw report text, \texttt{\{alias\_clause\}} optionally appends clinical synonyms (e.g., \textit{also known as splenic} for the spleen), and \texttt{\{description\}} is the organ-specific text extracted by the preceding stage. Full implementation details are provided in \texttt{preprocess\_reports.py} in the given code.
}

\label{fig:prompt_templates}
\label{qwen_prompt}
\end{figure*}

\begin{figure*}[t]
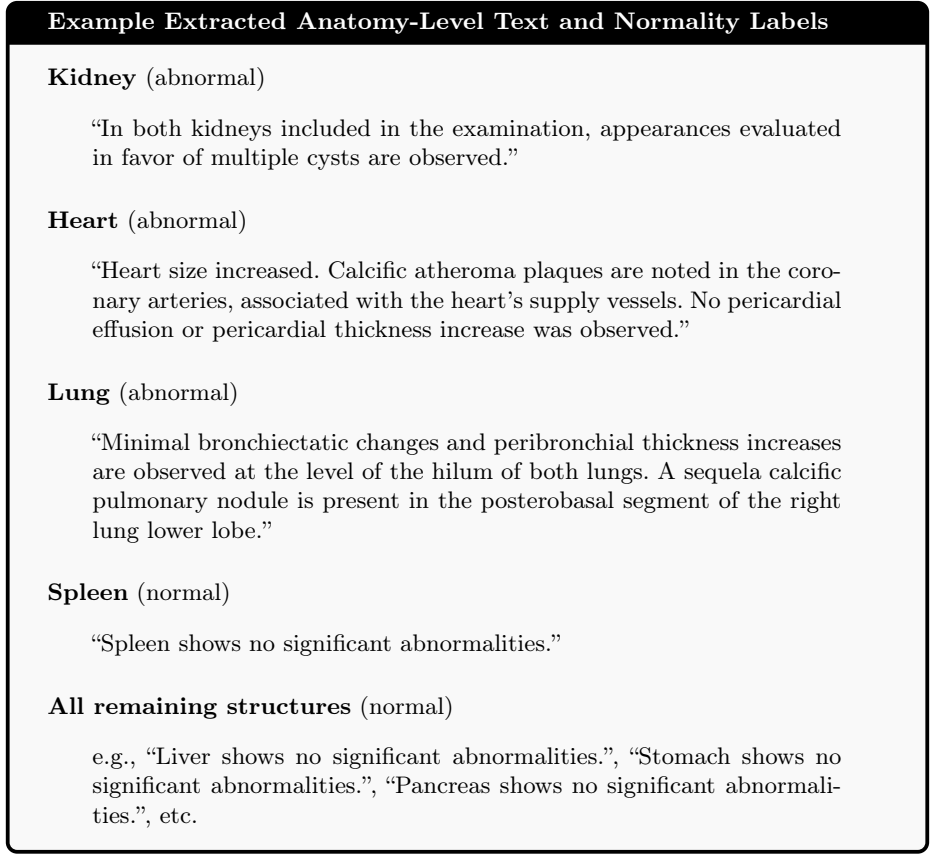

\centering
\begin{tcolorbox}[
    colback=gray!5,
    colframe=black,
    title={Example Extracted Anatomy-Level Text and Normality Labels},
    width=\textwidth,
    fonttitle=\bfseries
]
\textbf{Kidney} (abnormal)
\begin{quote}
\small ``In both kidneys included in the examination, appearances evaluated in favor of multiple cysts are observed.''
\end{quote}
\vspace{0.5em}
\textbf{Heart} (abnormal)
\begin{quote}
\small ``Heart size increased. Calcific atheroma plaques are noted in the coronary arteries, associated with the heart's supply vessels. No pericardial effusion or pericardial thickness increase was observed.''
\end{quote}
\vspace{0.5em}
\textbf{Lung} (abnormal)
\begin{quote}
\small ``Minimal bronchiectatic changes and peribronchial thickness increases are observed at the level of the hilum of both lungs. A sequela calcific pulmonary nodule is present in the posterobasal segment of the right lung lower lobe.''
\end{quote}
\vspace{0.5em}
\textbf{Spleen} (normal)
\begin{quote}
\small ``Spleen shows no significant abnormalities.''
\end{quote}
\vspace{0.5em}
\textbf{All remaining structures} (normal)
\begin{quote}
\small e.g., ``Liver shows no significant abnormalities.'', ``Stomach shows no significant abnormalities.'', ``Pancreas shows no significant abnormalities.'', etc.
\end{quote}
\end{tcolorbox}
\caption{Example anatomy-specific findings extracted from a radiology report using Qwen3-4B-Instruct, along with the corresponding binary normality label used in the soft-target contrastive loss described in Section~\ref{sec:loss}. Anatomical structures with no reported findings receive a default normal description.}
\label{fig:anatomy_text_examples}
\end{figure*}

\begin{figure*}[t]
\centering
\begin{tcolorbox}[
    colback=gray!5,
    colframe=black,
    title={Example Zero-Shot Finding Prompts},
    width=\textwidth,
    fonttitle=\bfseries
]

\textbf{Atelectasis}
\begin{itemize}
    \item \textit{Positive:} ``atelectasis'', ``bibasilar atelectasis'', ``subsegmental atelectasis'', ``lobar atelectasis''
    \item \textit{Negative:} ``no atelectasis'', ``lungs are clear'', ``no airspace opacities or atelectasis''
\end{itemize}

\vspace{0.5em}

\textbf{Cardiomegaly}
\begin{itemize}
    \item \textit{Positive:} ``cardiomegaly'', ``enlarged heart'', ``cardiac silhouette is enlarged'', ``moderate cardiomegaly''
    \item \textit{Negative:} ``no cardiomegaly'', ``heart is normal in size'', ``cardiac silhouette is normal''
\end{itemize}

\vspace{0.5em}

\textbf{Hiatal Hernia}
\begin{itemize}
    \item \textit{Positive:} ``hiatal hernia'', ``sliding hiatal hernia'', ``paraesophageal hernia''
    \item \textit{Negative:} ``no hiatal hernia'', ``hiatus is normal'', ``no herniation through the diaphragmatic hiatus''
\end{itemize}

\end{tcolorbox}

\caption{
Example positive and negative text prompts used for zero-shot finding classification. For each finding, a set of positive prompts describing the pathology and negative prompts describing normal appearance are encoded by the frozen text encoder and averaged to obtain class-level embeddings $f^{+}$ and $f^{-}$.
}
\label{fig:finding_prompts}
\end{figure*}

\section{Zero-Shot Scoring and Anatomy-Guided Evaluation}
\label{supp:anatomy_guided}

For all models, a scan-level score is computed by averaging projected anatomy embeddings across all present anatomical structures $\mathcal{P}$ and taking the difference in cosine similarity to each class:
\begin{equation}
    c_{\text{mean}} = \cos\!\left(
        \frac{1}{|\mathcal{P}|}\sum_{a \in \mathcal{P}} f_a^{\text{img}},\; f^{+}
    \right)
    - \cos\!\left(
        \frac{1}{|\mathcal{P}|}\sum_{a \in \mathcal{P}} f_a^{\text{img}},\; f^{-}
    \right)
\end{equation}

A finding is predicted as positive when $c_{\text{mean}}>0$. For ACA, we additionally evaluate an anatomy-guided scoring variant that exploits the per-anatomy structure of the model's representations. Rather than pooling across all present anatomical structures, the scan embedding for each finding $c$ is computed by restricting to an anatomically relevant subset $\mathcal{P}_c$:

\begin{equation}
    c_{\text{guided}} = \cos\!\left(
        \frac{1}{|\mathcal{P}_c|}\sum_{a \in \mathcal{P}_c} f_a^{\text{img}},\; f^{+}
    \right)
    - \cos\!\left(
        \frac{1}{|\mathcal{P}_c|}\sum_{a \in \mathcal{P}_c} f_a^{\text{img}},\; f^{-}
    \right)
\end{equation}

The mapping from finding to subset is contained in Table \ref{tab:anatomy_subsets}. For findings without a well-defined anatomy subset (e.g., lymphadenopathy, free air), $\mathcal{P}_c$ falls back to all present structures. The final anatomy-guided score is a linear blend with the mean-pool score $c_{\text{mean}}$:

\begin{equation}
    c = \alpha \cdot c_{\text{guided}} + (1 - \alpha) \cdot c_{\text{mean}}
\end{equation}

where $\alpha \in \{0.0, 0.2, 0.4, 0.5, 0.6, 0.8, 1.0\}$ is selected on the validation set by maximizing macro-average AUROC, and the same value is applied at test time for both \textbf{ACA (Anatomy-Guided)} and \textbf{ACA w/o $\mathcal{L}_{\text{scan}}$ (Anatomy-Guided)}.

\begin{table}[h]
\centering
\small
\resizebox{\textwidth}{!}{%
\begin{tabular}{ll}
\toprule
\textbf{Finding} & \textbf{Anatomy Subset} \\
\midrule
Atelectasis                    & Lung \\
Pleural Effusion               & Lung \\
Emphysema                      & Lung \\
Lung Nodule                    & Lung \\
Lung Opacity                   & Lung \\
Pulmonary Fibrotic Sequela     & Lung \\
Mosaic Attenuation Pattern     & Lung \\
Consolidation                  & Lung \\
Bronchiectasis                 & Lung \\
Interlobular Septal Thickening & Lung \\
Peribronchial Thickening       & Lung, Trachea \\
Cardiomegaly                   & Heart \\
Pericardial Effusion           & Heart \\
Coronary Calcification         & Heart \\
Arterial Wall Calcification    & Aorta, Iliac, Subclavian, \& Common Carotid Arteries \\
Coronary Artery Wall Calcification & Heart, Aorta \\
Aortic Valve Calcification     & Heart, Aorta \\
Abdominal Aortic Aneurysm      & Aorta \\
Atherosclerosis                & Aorta, Iliac Artery \\
Hiatal Hernia                  & Stomach, Esophagus \\
Hepatomegaly                   & Liver \\
Hepatic Steatosis              & Liver \\
Biliary Ductal Dilation        & Liver, Gallbladder \\
Gallstones                     & Gallbladder \\
Surgically Absent Gallbladder  & Gallbladder \\
Splenomegaly                   & Spleen \\
Renal Cyst                     & Kidney \\
Renal Hypodensities            & Kidney \\
Hydronephrosis                 & Kidney \\
Pancreatic Atrophy             & Pancreas \\
Prostatomegaly                 & Prostate \\
Submucosal Edema               & Small Bowel, Colon \\
Bowel Obstruction              & Small Bowel, Colon \\
Appendicitis                   & Colon \\
Thrombosis                     & Portal/Splenic Vein, Vena Cava, Iliac Vein \\
Metastatic Disease             & Liver, Lung, Rib, Lumbar Vertebrae, Thoracic Vertebrae \\
Osteopenia                     & Lumbar Vertebrae, Thoracic Vertebrae, Rib \\
Fracture                       & Rib, Lumbar Vertebrae, Thoracic Vertebrae \\
Ascites                        & Liver, Spleen \\
Anasarca & All present structures \\
Lymphadenopathy & All present structures \\
Free Air & All present structures \\
Medical Material & All present structures \\
\bottomrule
\end{tabular}
}
\caption{Anatomy subsets used for anatomy-guided zero-shot scoring per finding. Findings evaluated on both Merlin and CT-RATE datasets use the respective dataset's assignment.}
\label{tab:anatomy_subsets}
\end{table}

\section{Implementation Details}
\label{supp:hyperparams}

All models are trained with the same optimization setup. Table~\ref{tab:shared_hparams} lists the shared training hyperparameters, and Table~\ref{tab:arch_hparams} lists the architecture hyperparameters for each model. Where Merlin and CT-RATE differ due to their respective backbone output dimensions, both values are shown as Merlin / CT-RATE.

\begin{table}[h]
\centering
\small
\begin{tabular}{lc}
\toprule
\textbf{Hyperparameter} & \textbf{Value} \\
\midrule
Batch size            & 64 \\
Epochs                & 50 \\
Early stopping patience & 5 \\
Learning rate         & $1 \times 10^{-4}$ \\
Weight decay          & 0.05 \\
Warmup fraction       & 0.05 \\
Gradient clip         & 1.0 \\
Projection hidden dim & 512 \\
Projection output dim & 256 \\
Temperature init ($\tau$) & 0.07 \\
\bottomrule
\end{tabular}
\caption{Training hyperparameters shared across all models and both datasets.}
\label{tab:shared_hparams}
\end{table}

\begin{table}[h]
\centering
\small
\resizebox{\textwidth}{!}{%
\begin{tabular}{lll}
\toprule
\textbf{Model} & \textbf{Hyperparameter} & \textbf{Value} \\
\midrule
\multirow{1}{*}{MLP}
  & Visual input dim & 2048 / 512 \\
\midrule
\multirow{3}{*}{fVLM}
  & Multi-scale input dim      & 3840 / 1024 \\
  & Cross-attention heads      & 4 \\
  & Cross-attention dropout    & 0.1 \\
\midrule
\multirow{4}{*}{Spatial Transformer}
  & Spatial features             & 49 / 144 \\
  & Token input dim            & 2048 / 512 \\
  & Transformer hidden dim     & 1024 \\
  & Transformer layers         & 2 \\
  & Transformer heads          & 4 \\
  & Dropout                    & 0.1 \\
\midrule
\multirow{5}{*}{ACA w/o $\mathcal{L}_{\text{scan}}$, ACA w/o $\mathcal{L}_{\text{anatomy}}$, ACA}
  & Visual input dim           & 2048 / 512 \\
  & Transformer hidden dim     & 1024 \\
  & Positional encoding dim & 44 \\
  & Transformer layers         & 2 \\
  & Transformer heads          & 4 \\
  & Dropout                    & 0.1 \\
\midrule
\multirow{9}{*}{ViSD-Boost}
  & Multi-scale input dim      & 3840 / 1024 \\
  & Cross-attention heads      & 4 \\
  & Cross-attention dropout    & 0.1 \\
  & VQ-VAE codebook size       & 512 \\
  & VQ-VAE embedding dim       & 512 \\
  & VQ-VAE Transformer layers  & 1 \\
  & VQ-VAE Transformer heads   & 8 \\
  & VQ commitment cost         & 0.25 \\
\bottomrule
\end{tabular}
}
\caption{Architecture hyperparameters per model. Where values differ between the Merlin and CT-RATE backbones, both are shown as Merlin / CT-RATE.}
\label{tab:arch_hparams}
\end{table}

\begin{table}[h]
\centering
\caption{Merlin: positive finding counts per split (unlabeled excluded).}
\label{tab:merlin-findings}
\begin{tabular}{lrrr}
\toprule
Finding & Train & Val & Test \\
\midrule
Atelectasis & 2665 & 902 & 903 \\
Surgically Absent Gallbladder & 2515 & 805 & 843 \\
Atherosclerosis & 2493 & 778 & 879 \\
Pleural Effusion & 2484 & 794 & 834 \\
Renal Cyst & 1650 & 544 & 546 \\
Ascites & 1048 & 366 & 365 \\
Anasarca & 1020 & 318 & 352 \\
Hiatal Hernia & 886 & 280 & 312 \\
Hepatic Steatosis & 793 & 250 & 272 \\
Gallstones & 478 & 148 & 148 \\
Fracture & 470 & 175 & 177 \\
Pancreatic Atrophy & 453 & 118 & 161 \\
Osteopenia & 376 & 119 & 147 \\
Submucosal Edema & 359 & 146 & 144 \\
Cardiomegaly & 352 & 123 & 119 \\
Splenomegaly & 324 & 109 & 141 \\
Prostatomegaly & 306 & 115 & 81 \\
Renal Hypodensities & 302 & 122 & 122 \\
Hydronephrosis & 266 & 72 & 112 \\
Thrombosis & 247 & 80 & 77 \\
Bowel Obstruction & 211 & 64 & 72 \\
Aortic Valve Calcification & 206 & 61 & 72 \\
Coronary Calcification & 166 & 57 & 63 \\
Hepatomegaly & 165 & 44 & 57 \\
Biliary Ductal Dilation & 110 & 33 & 52 \\
Appendicitis & 106 & 32 & 43 \\
Lymphadenopathy & 104 & 28 & 49 \\
Free Air & 100 & 37 & 45 \\
Metastatic Disease & 83 & 24 & 46 \\
Abdominal Aortic Aneurysm & 71 & 16 & 17 \\
\bottomrule
\end{tabular}
\end{table}

\begin{table}[h]
\centering
\caption{CT-RATE: positive finding counts per split.}
\label{tab:ctrate-findings}
\begin{tabular}{lrrr}
\toprule
Finding & Train & Val & Test \\
\midrule
Lung nodule & 17032 & 4349 & 1361 \\
Lung opacity & 13845 & 3575 & 1184 \\
Arterial wall calcification & 10697 & 2679 & 867 \\
Pulmonary fibrotic sequela & 10127 & 2461 & 831 \\
Atelectasis & 9843 & 2418 & 713 \\
Lymphadenopathy & 9684 & 2534 & 789 \\
Coronary artery wall calcification & 9634 & 2390 & 765 \\
Emphysema & 7291 & 1831 & 600 \\
Consolidation & 6617 & 1701 & 581 \\
Hiatal hernia & 5414 & 1337 & 417 \\
Medical material & 4668 & 1150 & 313 \\
Pleural effusion & 4582 & 1122 & 376 \\
Cardiomegaly & 4259 & 1049 & 325 \\
Peribronchial thickening & 3960 & 1013 & 355 \\
Bronchiectasis & 3825 & 907 & 330 \\
Interlobular septal thickening & 2976 & 769 & 249 \\
Mosaic attenuation pattern & 2871 & 767 & 253 \\
Pericardial effusion & 2712 & 700 & 226 \\
\bottomrule
\end{tabular}
\end{table}

\begin{table*}[t]
\centering
\small
\begin{tabular}{lrr}
\toprule
\textbf{Finding} & \textbf{Positive} & \textbf{Negative} \\
\midrule
\multicolumn{3}{l}{\textbf{CT-RATE In-Distribution}} \\
\midrule
Medical Material & 313 & 2726 \\
Arterial Wall Calcification & 867 & 2172 \\
Cardiomegaly & 325 & 2714 \\
Pericardial Effusion & 226 & 2813 \\
Coronary Artery Wall Calcification & 765 & 2274 \\
Hiatal Hernia & 417 & 2622 \\
Lymphadenopathy & 789 & 2250 \\
Emphysema & 600 & 2439 \\
Atelectasis & 713 & 2326 \\
Lung Nodule & 1361 & 1678 \\
Lung Opacity & 1184 & 1855 \\
Pulmonary Fibrotic Sequela & 831 & 2208 \\
Pleural Effusion & 376 & 2663 \\
Mosaic Attenuation Pattern & 253 & 2786 \\
Peribronchial Thickening & 355 & 2684 \\
Consolidation & 581 & 2458 \\
Bronchiectasis & 330 & 2709 \\
Interlobular Septal Thickening & 249 & 2790 \\
\midrule
\multicolumn{3}{l}{\textbf{CT-RATE Out-of-Distribution}} \\
\midrule
Atelectasis & 713 & 2326 \\
Cardiomegaly & 325 & 2714 \\
Hiatal Hernia & 417 & 2622 \\
Lymphadenopathy & 789 & 2250 \\
Pleural Effusion & 376 & 2663 \\
Arterial Wall Calcification & 867 & 2172 \\
Coronary Artery Wall Calcification & 765 & 2274 \\
\bottomrule
\end{tabular}
\caption{Number of positive and sampled negative cases for each finding evaluated in the in-distribution and out-of-distribution settings on the CT-RATE dataset.}
\label{tab:finding_counts_ct_rate}
\end{table*}

\begin{table*}[t]
\centering
\small
\begin{tabular}{lrr}
\toprule
\textbf{Finding} & \textbf{Positive} & \textbf{Negative} \\
\midrule
\multicolumn{3}{l}{\textbf{Merlin In-Distribution}} \\
\midrule
Submucosal Edema & 144 & 162 \\
Renal Hypodensities & 122 & 1355 \\
Aortic Valve Calcification & 72 & 69 \\
Coronary Calcification & 63 & 372 \\
Thrombosis & 77 & 32 \\
Metastatic Disease & 46 & 123 \\
Pancreatic Atrophy & 161 & 3197 \\
Renal Cyst & 546 & 1412 \\
Osteopenia & 147 & 1069 \\
Surgically Absent Gallbladder & 843 & 2186 \\
Atelectasis & 903 & 990 \\
Abdominal Aortic Aneurysm & 17 & 79 \\
Anasarca & 352 & 1992 \\
Hiatal Hernia & 312 & 1847 \\
Lymphadenopathy & 49 & 205 \\
Prostatomegaly & 81 & 1213 \\
Biliary Ductal Dilation & 52 & 211 \\
Cardiomegaly & 119 & 291 \\
Splenomegaly & 141 & 3225 \\
Hepatomegaly & 57 & 1150 \\
Atherosclerosis & 879 & 55 \\
Ascites & 365 & 183 \\
Pleural Effusion & 834 & 556 \\
Hepatic Steatosis & 272 & 27 \\
Appendicitis & 43 & 52 \\
Gallstones & 148 & 248 \\
Hydronephrosis & 112 & 2079 \\
Bowel Obstruction & 72 & 1345 \\
Free Air & 45 & 929 \\
Fracture & 177 & 196 \\
\midrule
\multicolumn{3}{l}{\textbf{Merlin Out-of-Distribution}} \\
\midrule
Coronary Calcification & 63 & 372 \\
Atelectasis & 903 & 990 \\
Hiatal Hernia & 312 & 1847 \\
Lymphadenopathy & 49 & 205 \\
Cardiomegaly & 119 & 291 \\
Atherosclerosis & 879 & 55 \\
Pleural Effusion & 834 & 556 \\
\bottomrule
\end{tabular}
\caption{Number of positive and sampled negative cases for each finding evaluated in the in-distribution and out-of-distribution settings on the Merlin dataset.}
\label{tab:finding_counts_merlin}
\end{table*}

\begin{table*}[t]
\centering
\small
\caption{Per-finding AUROC on the Merlin in-distribution test set. Macro average is computed over all 30 findings. Overlapping Macro average restricts the macro-average to the 7 findings shared with CT-RATE, matching the finding set used for out-of-distribution evaluation (Table~\ref{tab:merlin_ood}). }
\label{tab:merlin_id}
\resizebox{\textwidth}{!}{
\begin{tabular}{lcccccc}
\toprule
Finding & Merlin & Spatial Transformer & MLP & fVLM & ViSD-Boost & ACA \\
\midrule
Submucosal edema & 0.7415 & 0.7309 & 0.7220 & 0.6832 & 0.6923 & 0.7185\\
Renal hypodensities & 0.6770 & 0.6731 & 0.7218 & 0.7081 & 0.7405 & 0.7512\\
Aortic valve calcification & 0.7997 & 0.8133 & 0.9506 & 0.9578 & 0.9557 & 0.9637\\
Coronary calcification & 0.8057 & 0.8198 & 0.8392 & 0.8442 & 0.8469 & 0.8496\\
Thrombosis & 0.6269 & 0.5759 & 0.5481 & 0.6041 & 0.6057 & 0.6175\\
Metastatic disease & 0.7186 & 0.7363 & 0.8305 & 0.8801 & 0.8643 & 0.8719\\
Pancreatic atrophy & 0.7180 & 0.7394 & 0.7744 & 0.7861 & 0.7706 & 0.7653\\
Renal cyst & 0.6204 & 0.6281 & 0.7436 & 0.6704 & 0.6754 & 0.7795\\
Osteopenia & 0.7435 & 0.7932 & 0.9199 & 0.9146 & 0.8990 & 0.9279\\
Surgically absent gallbladder & 0.9741 & 0.9688 & 0.9214 & 0.7125 & 0.8299 & 0.9794\\
Atelectasis & 0.6757 & 0.7256 & 0.7800 & 0.6530 & 0.7567 & 0.8189\\
Abdominal aortic aneurysm & 0.7501 & 0.8679 & 0.7791 & 0.7732 & 0.7492 & 0.7826\\
Anasarca & 0.9279 & 0.9140 & 0.9002 & 0.9303 & 0.9352 & 0.9378\\
Hiatal hernia & 0.6334 & 0.6797 & 0.7721 & 0.7625 & 0.7295 & 0.7604\\
Lymphadenopathy & 0.7870 & 0.7291 & 0.6910 & 0.6541 & 0.7131 & 0.7469\\
Prostatomegaly & 0.6765 & 0.7324 & 0.7535 & 0.7425 & 0.7315 & 0.7658\\
Biliary ductal dilation & 0.7908 & 0.8189 & 0.7831 & 0.7898 & 0.8044 & 0.8181\\
Cardiomegaly & 0.8247 & 0.8545 & 0.8801 & 0.8458 & 0.8663 & 0.8925\\
Splenomegaly & 0.9012 & 0.8836 & 0.9120 & 0.9058 & 0.9197 & 0.9267\\
Hepatomegaly & 0.7645 & 0.7486 & 0.7799 & 0.8221 & 0.8318 & 0.8672\\
Atherosclerosis & 0.9621 & 0.9602 & 0.8683 & 0.8382 & 0.8808 & 0.8899\\
Ascites & 0.9022 & 0.9114 & 0.8711 & 0.8351 & 0.8532 & 0.9310\\
Pleural effusion & 0.9294 & 0.9246 & 0.8691 & 0.8525 & 0.9204 & 0.9455\\
Hepatic steatosis & 0.6747 & 0.7628 & 0.7698 & 0.7809 & 0.7192 & 0.6580\\
Appendicitis & 0.7185 & 0.7123 & 0.7333 & 0.5840 & 0.5012 & 0.7344\\
Gallstones & 0.7566 & 0.7221 & 0.6920 & 0.6212 & 0.7251 & 0.7374\\
Hydronephrosis & 0.7295 & 0.7304 & 0.7313 & 0.7611 & 0.7499 & 0.7655\\
Bowel obstruction & 0.8668 & 0.8888 & 0.8868 & 0.8489 & 0.7493 & 0.8791\\
Free air & 0.7648 & 0.7802 & 0.8249 & 0.7487 & 0.7294 & 0.7961\\
Fracture & 0.7246 & 0.6947 & 0.6941 & 0.5871 & 0.6622 & 0.7615\\
\midrule
Overlapping Macro average & 0.8026 & 0.8134 & 0.8143 & 0.7786 & 0.8163 & 0.8434\\
\midrule
Macro average & 0.7729 & 0.7840 & 0.7981 & 0.7699 & 0.7803 & 0.8213\\

\bottomrule
\end{tabular}}
\end{table*}

\begin{table*}[t]
\centering
\small
\caption{Per-finding AUROC on the CT-RATE in-distribution test set. Macro average is computed over all 18 findings. Overlapping Macro average restricts the macro-average to the 7 findings shared with Merlin, matching the finding set used for out-of-distribution evaluation (Table~\ref{tab:ctrate_ood}). }
\label{tab:ctrate_id}
\resizebox{\textwidth}{!}{
\begin{tabular}{lcccccc}
\toprule
Finding & CT-CLIP & Spatial Transformer & MLP & fVLM & ViSD-Boost & ACA \\
\midrule
Medical material & 0.6462 & 0.6208 & 0.6252 & 0.6866 & 0.6784 & 0.6798 \\
Arterial wall calcification & 0.8492 & 0.7892 & 0.6513 & 0.8135 & 0.7810 & 0.8768 \\
Cardiomegaly & 0.8686 & 0.7745 & 0.6677 & 0.7815 & 0.8420 & 0.8973 \\
Pericardial effusion & 0.7053 & 0.6543 & 0.6337 & 0.7394 & 0.7692 & 0.7787 \\
Coronary artery wall calcification & 0.8377 & 0.7655 & 0.7052 & 0.7953 & 0.7748 & 0.8505 \\
Hiatal hernia & 0.7187 & 0.6843 & 0.6466 & 0.6532 & 0.6605 & 0.7956 \\
Lymphadenopathy & 0.6807 & 0.6226 & 0.5633 & 0.6444 & 0.5585 & 0.6543 \\
Emphysema & 0.7065 & 0.6256 & 0.5965 & 0.6739 & 0.6715 & 0.7506 \\
Atelectasis & 0.6449 & 0.6220 & 0.6058 & 0.6567 & 0.6098 & 0.6679 \\
Lung nodule & 0.5381 & 0.4701 & 0.5473 & 0.5283 & 0.4869 & 0.5918 \\
Lung opacity & 0.6743 & 0.5990 & 0.5440 & 0.6095 & 0.5814 & 0.6078 \\
Pulmonary fibrotic sequela & 0.5713 & 0.5197 & 0.5351 & 0.5651 & 0.5246 & 0.6230 \\
Pleural effusion & 0.8952 & 0.8306 & 0.6624 & 0.8485 & 0.7648 & 0.8744 \\
Mosaic attenuation pattern & 0.7470 & 0.7179 & 0.6584 & 0.7176 & 0.6818 & 0.7519 \\
Peribronchial thickening & 0.6079 & 0.5412 & 0.6506 & 0.6678 & 0.5395 & 0.6735 \\
Consolidation & 0.6992 & 0.6249 & 0.5302 & 0.6291 & 0.6035 & 0.6442 \\
Bronchiectasis & 0.6560 & 0.5675 & 0.5815 & 0.5871 & 0.5949 & 0.7182 \\
Interlobular septal thickening & 0.7005 & 0.6100 & 0.6279 & 0.7180 & 0.5973 & 0.7245 \\
\midrule
Overlapping Macro average & 0.7850 & 0.7270 & 0.6432 & 0.7419 & 0.7131 & 0.8024\\
\midrule
Macro average & 0.7082 & 0.6467 & 0.6129 & 0.6842 & 0.6511 & 0.7311 \\
\bottomrule
\end{tabular}}
\end{table*}

\begin{table}[t]
\centering
\small
\caption{Per-finding AUROC on the Merlin out-of-distribution test set.}
\label{tab:merlin_ood}
\resizebox{\textwidth}{!}{
\begin{tabular}{lcccccc}
\toprule
Finding & CT-CLIP & Spatial Transformer & MLP & fVLM & ViSD-Boost & ACA\\
\midrule
Atelectasis & 0.5784 & 0.5903 & 0.5871 & 0.5763 & 0.7383 & 0.6939\\
Cardiomegaly & 0.6443 & 0.5604 & 0.5535 & 0.5456 & 0.5789 & 0.6598\\
Hiatal hernia & 0.5995 & 0.6398 & 0.6168 & 0.5795 & 0.6553 & 0.6822\\
Lymphadenopathy & 0.4493 & 0.4750 & 0.4812 & 0.5282 & 0.5067 & 0.5687\\
Pleural effusion & 0.5838 & 0.6409 & 0.6951 & 0.6421 & 0.6765 & 0.7691\\
Atherosclerosis & 0.5662 & 0.4954 & 0.5579 & 0.5624 & 0.6448 & 0.6438\\
Coronary calcification & 0.6541 & 0.5220 & 0.4974 & 0.4770 & 0.4968 & 0.6886\\
\midrule
Macro average & 0.5822 & 0.5605 & 0.5699 & 0.5587 & 0.6139 & 0.6723\\
\bottomrule
\end{tabular}
}
\end{table}

\begin{table}[t]
\centering
\small
\caption{Per-finding AUROC on the CT-RATE out-of-distribution test set.}
\label{tab:ctrate_ood}
\resizebox{\textwidth}{!}{
\begin{tabular}{lcccccc}
\toprule
Finding & Merlin & Spatial Transformer & MLP & fVLM & ViSD-Boost & ACA\\
\midrule
Atelectasis & 0.5753 & 0.5856 & 0.5845 & 0.4770 & 0.5951 & 0.6050\\
Cardiomegaly & 0.7875 & 0.8044 & 0.8519 & 0.8265 & 0.8583 & 0.8448\\
Hiatal hernia & 0.6120 & 0.6192 & 0.6711 & 0.6766 & 0.6487 & 0.6565\\
Lymphadenopathy & 0.5873 & 0.5927 & 0.5350 & 0.5166 & 0.5940 & 0.6119\\
Pleural effusion & 0.9002 & 0.9118 & 0.8462 & 0.8062 & 0.9024 & 0.8789\\
Arterial wall calcification & 0.6853 & 0.7014 & 0.8409 & 0.8042 & 0.7880 & 0.7853\\
Coronary artery wall calcification & 0.6959 & 0.6871 & 0.8424 & 0.8178 & 0.8024 & 0.7979\\
\midrule
Macro average & 0.6919 & 0.7003 & 0.7389 & 0.7036 & 0.7413 & 0.7400\\
\bottomrule
\end{tabular}
}
\end{table}

\end{document}